\documentclass{IEEEtran}
\IEEEoverridecommandlockouts
\usepackage[space]{cite}
\usepackage{authblk}
\usepackage{amsmath, amsthm, amssymb, amsfonts, bm}
\usepackage{mathtools}
\usepackage{tikz}
\usepackage[skip = 0pt]{subcaption}
\newtheorem{theorem}{Theorem}
\newtheorem{lemma}[theorem]{Lemma}

\newtheorem{corollary}[theorem]{Corollary}

\newtheorem{prop}[theorem]{Proposition}
\newtheorem{deff}[theorem]{Definition}
\newtheorem{assumption}{Assumption}
\usepackage{algorithm}
\usepackage{algorithmic}

\usepackage{array}
\usepackage{graphicx}
\usepackage{textcomp}
\usepackage{xcolor}
\usepackage{adjustbox}
\usepackage{fancyhdr}
\definecolor{peachfill}{RGB}{255, 230, 204}
\definecolor{darkorangeborder}{RGB}{215, 155, 0}
\definecolor{pinkfill}{RGB}{248, 206, 204}
\definecolor{darkredborder}{RGB}{184, 84, 80}
\definecolor{bluefill}{RGB}{218, 232, 252}
\definecolor{blueborder}{RGB}{108, 142, 191}
\definecolor{greenfill}{RGB}{213, 232, 212}
\definecolor{greenborder}{RGB}{130, 179, 102}
\definecolor{purplefill}{RGB}{225, 213, 231}
\definecolor{purpleborder}{RGB}{150, 115, 166}
\definecolor{yellowfill}{RGB}{255, 242, 204}
\definecolor{yellowborder}{RGB}{214, 182, 86}
\DeclareRobustCommand{\rchi}{{\mathpalette\irchi\relax}}
\newcommand{\irchi}[2]{\raisebox{\depth}{$#1\chi$}}
\newcommand{\norm}[1]{\lVert#1\rVert}
\newcommand\middlescript[1]{\vcenter{\hbox{$\scriptstyle #1$}}}
\usepackage{nomencl}
\makenomenclature

\usepackage{hyperref}
\hypersetup{
    colorlinks=true,
    linkcolor=blue,
    filecolor=magenta,      
    urlcolor=cyan,
    pdftitle={Overleaf Example},
    pdfpagemode=FullScreen,
    }

\fancypagestyle{firststyle}
{
   \fancyhf{}
   \fancyfoot[C]{\footnotesize This work has been submitted to the IEEE for possible publication. Copyright may be transferred without notice, after which this version may no longer be accessible.}
}

\begin{document}
\bstctlcite{IEEEexample:BSTcontrol}
\title{On Generalization Bounds for Deep Compound Gaussian Neural Networks
}
\author{{\large Carter Lyons, Raghu G. Raj, and Margaret Cheney}
\thanks{Carter Lyons is with the U.S. Naval Research Laboratory, Washington, DC 20375 USA and also with Colorado State University, Fort Collins, CO 80523 USA (e-mail: carter.lyons@colostate.edu)}
\thanks{Raghu G. Raj is with the U.S. Naval Research Laboratory, Washington, DC 20375 USA (e-mail: raghu.g.raj@ieee.org)}
\thanks{Margaret Cheney is with Colorado State University, Fort Collins, CO 80523 USA (e-mail: margaret.cheney@colostate.edu)}
\thanks{This work was sponsored in part by the Office of Naval Research via the NRL base program and under award numbers N00014-21-1-2145 and N0001423WX01875. Furthermore, this work was sponsored in part by the Air Force Office of Scientific Research under award number FA9550-21-1-0169.}
}

\maketitle
\thispagestyle{firststyle}

\begin{abstract}
Algorithm unfolding or unrolling is the technique of constructing a deep neural network (DNN) from an iterative algorithm. Unrolled DNNs often provide better interpretability and superior empirical performance over standard DNNs in signal estimation tasks. An important theoretical question, which has only recently received attention, is the development of generalization error bounds for unrolled DNNs. These bounds deliver theoretical and practical insights into the performance of a DNN on empirical datasets that are distinct from, but sampled from, the probability density generating the DNN training data. In this paper, we develop novel generalization error bounds for a class of unrolled DNNs that are informed by a compound Gaussian prior. These compound Gaussian networks have been shown to outperform comparative standard and unfolded deep neural networks in compressive sensing and tomographic imaging problems. The generalization error bound is formulated by bounding the Rademacher complexity of the class of compound Gaussian network estimates with Dudley's integral. Under realistic conditions, we show that, at worst, the generalization error scales $\mathcal{O}(n\sqrt{\ln(n)})$ in the signal dimension and $\mathcal{O}(($Network Size$)^{3/2})$ in network size.
\end{abstract}


\nomenclature[01]{$\mathbb{R}$}{Set of real numbers.}
\nomenclature[02]{$\bm{v}$}{$=[v_i]\in\mathbb{R}^d$. Boldface characters are vectors.}
\nomenclature[03]{$(\cdot)^T$}{Transpose of vector or matrix $(\cdot)$.}
\nomenclature[04]{$\odot$}{Hadamard product.}
\nomenclature[05]{$f(\bm{v})$}{$=[f(v_i)]$ for componentwise function $f:\mathbb{R}\to\mathbb{R}$.}
\nomenclature[06]{$\mathbb{N}[d]$}{$= \{1, 2, \ldots, d\}$ is set of first $d$ natural numbers.}
\nomenclature[07]{$\scalebox{.92}{$\mathcal{P}$}_{a,b}\scalebox{.85}{$(x)$}$}{$=a+\textnormal{ReLU}(x-a)-\textnormal{ReLU}(x-b)$, for $a, b\in\mathbb{R}$, is a modified ReLU (mReLU) activation function.}
\nomenclature[08]{$\rho_{b}(\bm{v})$}{$=\bm{v}/\max\{1,b^{-1}\norm{\bm{v}}_2\}$ for $b > 0$, is the projection onto the Euclidean ball of radius $b$.}
\nomenclature[09]{$A$}{$\in\mathbb{R}^{m\times n}$ is a measurement, observation, or sensing matrix.}
\nomenclature[10]{$A_{\bm{z}}$}{$= A\textnormal{Diag}(\bm{z})\in\mathbb{R}^{m\times n}$ for any vector $\bm{z}\in\mathbb{R}^n$.}
\nomenclature[11]{$\mathcal{T}_{\bm{y}}(\bm{z})$}{$\equiv \mathcal{T}_{\bm{y}}(\bm{z}; P_u) \coloneqq (A_{\bm{z}}^TA_{\bm{z}} + P_u^{-1})^{-1}A_{\bm{z}}^T\bm{y}$.}
\nomenclature[12]{$\mathcal{H}^{(i)}_{\textnormal{CG}}$}{Hypothesis class for G-CG-Net, CG-Net, and DR-CG-Net when $i = 1$, $i = 2$, and $i = 3$, respectively.}
\printnomenclature

\section{Introduction}

\IEEEPARstart{M}{achine} learning success in image classification has recently spurred its application, in particular with deep neural networks (DNN), in signal estimation tasks. Signal estimation is one example of an inverse problem where we desire a reconstructed signal from some undersampled measurements. Particular applications of interest include X-Ray computed tomography (CT), magnetic resonance imaging (MRI), and compressive sensing.

While DNNs have been shown to outperform iterative approaches in estimated signal quality and computational time~\cite{learned_ISTA, algorithm_unrolling, zhang2018ista, MADUN, CGNetTSP, DRCGNetTCI}, this typically requires a significant amount of training data and training time. Certain problems of interest, such as MRI, do not have large training datasets readily available and thus DNN performance can falter on these problems. Additionally, standard DNNs, e.g. convolutional neural networks, do not incorporate any outside prior information into the estimation model as is included in iterative approaches.

Algorithm unfolding or unrolling is a technique, introduced by Gregor and LeCun~\cite{learned_ISTA}, which combines the iterative approaches and deep-learning methods by structuring the layers of a DNN such that they correspond to an iteration from the iterative algorithm. While a standard DNN acts as a black-box process, we have an understanding of the inner workings of an unrolled DNN from understanding the original iterative algorithm. Furthermore, algorithm unrolling allows for the incorporation of prior information into the deep-learning framework. Example iterative algorithms that have been unrolled into DNNs include: iterative shrinkage and thresholding algorithm (ISTA)~\cite{MADUN, song2023MAPUN, zhang2018ista, learned_ISTA, learned_ISTA_xray, you2021ista, xiang2021FISTANet} and proximal gradient descent~\cite{learned_proximal_operators, deep_priors}. These unrolled DNNs have shown excellent performance in image estimation while offering simple interpretability of the network layers~\cite{algorithm_unrolling}. However, these unrolled DNNs still require large training datasets.

Recently, two compound Gaussian (CG) informed DNNs were developed through the use of algorithm unrolling. These unrolled CG-based DNNs have empirically produced superior estimated signals over comparative iterative and DNN methods, especially in scenarios of low training data~\cite{APSIPAlyonsrajcheney, Asilomarlyonsrajcheney, CGNetTSP, DRCGNetTCI}. While this success was shown empirically, no generalization guarantees have been provided, a gap that this paper fills. Specifically, we:
\begin{enumerate}
    \item Establish and prove a Lipschitz property of the outputs from unrolled, CG-informed DNNs with respect to (w.r.t.) the DNN parameters.
    \item Develop an encompassing generalization error bound (GEB) for unrolled, CG-informed DNNs by bounding Rademacher complexities with covering numbers.
    \item Apply the developed GEB and provide asymptotic forms in terms of DNN size and signal dimension for two distinct formulations of unrolled, CG-based DNNs named compound Gaussian network (CG-Net)~\cite{CGNetTSP} and deep regularized compound Gaussian network (DR-CG-Net)~\cite{DRCGNetTCI}. We theoretically demonstrate that DR-CG-Net exhibits a tighter GEB than CG-Net, which was empirically observed in~\cite{CGNetTSP, DRCGNetTCI}.
\end{enumerate}

The GEB we develop is formulated using techniques informed by those discussed in~\cite{NNfromstatistical}, which produces a GEB for an ISTA unrolled DNN with learned sparsity transformation. Such techniques are similarly used in~\cite{kouni2022deconet} for an unrolled DNN employing a learned analysis sparsity transformation and in~\cite{joukovsky2021generalization} for a $\ell_1$-$\ell_1$ unrolled recurrent neural network.

\subsection{Generalization Error}

A DNN can be viewed as a collection of ordered layers, denoted $\bm{L}_0, \bm{L}_1, \ldots, \bm{L}_K$ for $K > 1$, where layers feed into one another from the input layer, $\bm{L}_0$, to the output layer, $\bm{L}_K$. Intermediate layers $\bm{L}_1, \ldots, \bm{L}_{K-1}$ are known as hidden layers. Each layer $\bm{L}_k$ contains $d_k$ hidden units~\cite{goodfellow2016deep} that are assigned a computed value when transmitting a signal through the DNN. 

A function, $\bm{f}_k:\mathbb{R}^{d_{i_1(k)}}\times \cdots \times \mathbb{R}^{d_{i_j(k)}} \to \mathbb{R}^{d_k}$, that is parameterized by some $\bm{\theta}_k$ defines the computation, i.e. signal transmission, at layer $\bm{L}_k$ where $\mathcal{I}_k\coloneqq\{i_1(k),\ldots, i_j(k)\}\subseteq \{0,1,\ldots, K-1\}$ are the indices of layers that feed into $\bm{L}_k$. That is, given an input signal, $\overline{\bm{y}}\in\mathbb{R}^{d_0}$, assigned to $\bm{L}_0$, a DNN is the composition of parameterized vector input and vector output functions where
 \[
\bm{L}_k \equiv \bm{f}_k\left(\bm{L}_{i_1(k)}, \ldots, \bm{L}_{i_j(k)}; \bm{\theta}_k\right).
\]

Let $\mathcal{C}$ and $\mathcal{Y}$ be the set of possible signals of interest, e.g. images, and possible signal measurements, e.g. image measurements, respectively. Further, let $(\mathcal{C}, \mathcal{A}_c, \mathcal{D}_c)$ be a probability space where $\mathcal{A}_c$ and $\mathcal{D}_c$ are a $\sigma$-algebra and unknown probability density on $\mathcal{C},$ respectively. Similarly, define the probability space $(\mathcal{Y}, \mathcal{A}_y, \mathcal{D}_y)$. Now, consider the probability space $(\mathcal{Y}\times \mathcal{C}, \mathcal{A}_y\otimes \mathcal{A}_c, \mathcal{D})$ where $\mathcal{D}$ is an unknown joint probability density with marginal distributions $\mathcal{D}_c$ and $\mathcal{D}_y$. Let $\mathcal{S} = \{(\overline{\bm{y}}_i, \overline{\bm{c}}_i)\}_{i\in \mathbb{N}[N_s]}$ be a training dataset where each pair $(\overline{\bm{y}}_i, \overline{\bm{c}}_i)$ is drawn i.i.d. from $\mathcal{D}$.
We denote the network parameters as $\bm{\Theta} = (\bm{\theta}_1, \ldots, \bm{\theta}_K)$, and let $\Omega_k$ be the space of $\bm{\theta}_k$ that can be learned by the DNN. Further, we write the network output, i.e. signal at $\bm{L}_K$, that is dependent on network parameters and input measurements $\bm{y}$, as $\widehat{\bm{c}}(\bm{y}; \bm{\Theta})$.  

The \textbf{hypothesis space}, $\mathcal{H}$, of a DNN is given by
\begin{align*}
    \mathcal{H} \coloneqq \{\widehat{\bm{c}}(\boldsymbol{\cdot}; \bm{\Theta}): \bm{\Theta}\in \Omega_1\times \cdots\times \Omega_K\}.
\end{align*}
For a loss function, $L\left(\bm{x}_1, \bm{x}_2\right)$, that measures a discrepancy between $\bm{x}_1\in\mathbb{R}^{d_K}$ and $\bm{x}_2\in\mathbb{R}^{d_K}$, the \textbf{empirical loss} of a hypothesis $\widehat{\bm{c}}\in\mathcal{H}$, over a training dataset $\mathcal{S}$, is given as
\begin{align*}
    \mathcal{L}_{\mathcal{S}}(\widehat{\bm{c}}) = \frac{1}{N_s}\sum_{i = 1}^{N_s} L\left(\widehat{\bm{c}}(\overline{\bm{y}}_i; \bm{\Theta}), \overline{\bm{c}}_i\right).
\end{align*}
Mean-squared error or mean-absolute error are common loss functions~\cite{NN_loss}. The \textbf{actual loss} of a hypothesis $\widehat{\bm{c}}\in\mathcal{H}$ is
\begin{align*}
    \mathcal{L}(\widehat{\bm{c}}) = \mathbb{E}_{(\bm{y}, \bm{c})\sim \mathcal{D}}\left[L(\widehat{\bm{c}}(\bm{y};\bm{\Theta}),\bm{c})\right].
\end{align*}
Finally, the \textbf{generalization error} (GE$_{\mathcal{S}}$) of a hypothesis $\widehat{\bm{c}}\in\mathcal{H}$ is the difference in the empirical and actual loss. That is,
\begin{align*}
    \textnormal{GE}_{\mathcal{S}}(\widehat{\bm{c}}) = |\mathcal{L}(\widehat{\bm{c}}) - \mathcal{L}_{\mathcal{S}}(\widehat{\bm{c}})|.
\end{align*}

A DNN learns its parameters $\bm{\Theta}$ by minimizing the empirical loss, i.e. by choosing a hypothesis $\widehat{\bm{c}}\in\mathcal{H}$ that minimizes $\mathcal{L}_{\mathcal{S}}(\widehat{\bm{c}})$. It is critical in applications, however, for a DNN to similarly generate excellent results when provided any new data sample drawn from $\mathcal{D}$ that is not contained in the training dataset. This is tantamount to minimizing the generalization error, and thus, obtaining an estimate or bound on GE$_{\mathcal{S}}$ for a DNN is of significant interest.

\section{Generalized Compound Gaussian Network}

The generalized compound Gaussian network (G-CG-Net) shown in Fig. \ref{fig:DR-CG-Net structure} is an unrolled, CG-based DNN for solving linear inverse problems. Note, the use of ``generalized" in the naming of G-CG-Net denotes the fact that this network encompasses the compound Gaussian network (CG-Net)~\cite{APSIPAlyonsrajcheney, CGNetTSP} and deep regularized compound Gaussian network (DR-CG-Net)~\cite{DRCGNetTCI, Asilomarlyonsrajcheney} as special cases. In contrast, the use of ``generalized" in terms of network error denotes a DNNs ability to transfer from training data to testing data.

Through the study of image statistics, it has been shown that sparsity coefficients of natural images exhibit self-similarity, heavy-tailed marginal distributions, and self-reinforcement among local coefficients~\cite{Wavelet_Trees}. These properties are encompassed by the class of CG densities~\cite{Scale_Mixtures, Wavelet_Trees, compound_gaussian}. Thus, the CG prior better captures statistical properties of natural images as well as images from other modalities such as radar~\cite{chance2011information, waveform_opt}. A useful formulation of the CG prior is modeling a signal as the Hadamard product $\bm{c} = \bm{z}\odot\bm{u}$ such that $\bm{u}\sim \mathcal{N}(\bm{0},\Sigma_u)$, $\bm{z}\sim p_{\bm{z}}$, and $\bm{u}$ and $\bm{z}$ are independent random variables~\cite{Wavelet_Trees,HB-MAP}. We call $\bm{z}$ the scale variable and $\bm{u}$ the Gaussian variable. The linear measurement model we consider is
\begin{equation}
    \bm{y} = A\bm{c} + \bm{\nu} \equiv A(\bm{z}\odot\bm{u}) + \bm{\nu}. \label{eqn:linear_msrmt}
\end{equation}
G-CG-Net is a method that recovers $\bm{c}$, by estimating $\bm{z}$ and $\bm{u}$, when given $\bm{y}$ and $A$.

\subsection{Iterative Algorithm}

Algorithm \ref{alg:CG-LS} provides pseudocode for the iterative algorithm generalized compound Gaussian least squares (G-CG-LS) to be unrolled into G-CG-Net. Consider the cost function
\begin{align}
    F(\bm{u},\bm{z}) = \frac{1}{2}\norm{\bm{y}-A(\bm{z}\odot\bm{u})}_2^2 + \frac{1}{2}\bm{u}^T P_u^{-1}\bm{u} + \mathcal{R}(\bm{z}). \label{eqn:cost function}
\end{align}
A maximum-a-posteriori (MAP) estimate of $\bm{z}$ and $\bm{u}$ from (\ref{eqn:linear_msrmt}) is a special case of (\ref{eqn:cost function}) 
when $P_u\propto \Sigma_u$ and $\mathcal{R}\propto \log(p_{\bm{z}}(\bm{z}))$.

Using block coordinate descent~\cite{wright2015coordinate}, G-CG-LS alternatively minimizes (\ref{eqn:cost function}) over $\bm{z}$ and $\bm{u}$. The minimum of (\ref{eqn:cost function}) in $\bm{u}$ is a Tikhonov solution, $\mathcal{T}_{\bm{y}}(\bm{z}) \equiv \mathcal{T}_{\bm{y}}(\bm{z}; P_u)$, given by
\begin{align*}
   \mathcal{T}_{\bm{y}}(\bm{z}) &\coloneqq  (A_{\bm{z}}^TA_{\bm{z}} + P_u^{-1})^{-1}A_{\bm{z}}^T\bm{y} \\
   &= P_u A_{\bm{z}}^T(I+A_{\bm{z}}P_uA_{\bm{z}}^T)^{-1}\bm{y}
\end{align*}
where the second equality results from using the Woodbury matrix identity.

\begin{algorithm}[t]
\caption{Generalized Compound Gaussian Least Squares}\label{alg:CG-LS}
\begin{algorithmic}[1]
\REQUIRE{Measurement $\bm{y}$. Initial estimate $\bm{z}_1^{(0)}$.} 
\STATE Choose a $\bm{u}_0$ (e.g. $\bm{u}_0 = \mathcal{T}_{\bm{y}}(\bm{z}_1^{(0)})$) \label{line:initial}
\FOR{$k\in \{1,2,\ldots, K\}$}
\STATE \underline{$\bm{z}$ \textsc{estimation}}:
\FOR{$j\in \{1,2, \ldots, J\}$}
\STATE  $\bm{z}_{k}^{(j)} =g(\bm{z}_{k}^{(j-1)}, \bm{u}_{k-1}) \equiv g(\bm{z}_{k}^{(j-1)}, \bm{u}_{k-1}; \bm{y}) $ \label{line:z update}
\ENDFOR
\STATE $\bm{z}_{k+1}^{(0)} = \bm{z}_k^{(J)}$
\STATE \underline{$\bm{u}$ \textsc{estimation}}:
\STATE $ \bm{u}_{k} = \mathcal{T}_{\bm{y}}(\bm{z}_k^{(J)})$ \label{line:u update}
\ENDFOR
\ENSURE{$\bm{c}^* = \bm{z}_K^{(J)}\odot \bm{u}_K$}
\end{algorithmic}
\end{algorithm}

As the minimum of (\ref{eqn:cost function}) in $\bm{z}$ does not have a closed form solution,  for general regularization $\mathcal{R}(\bm{z})$, we iteratively minimize by performing $J$ descent steps on $\bm{z}$. That is, for $g:\mathbb{R}^n\times\mathbb{R}^n \to\mathbb{R}^n$ a scale-variable-descent update, the estimate of $\bm{z}$ on descent step $j$ of iteration $k$, denoted as $\bm{z}_k^{(j)}$, is given by $\bm{z}_k^{(j)} = g(\bm{z}_k^{(j-1)}, \bm{u}_{k-1}; \bm{y})$ where $\bm{z}_{k+1}^{(0)} = \bm{z}_k^{(J)}$. For instance, \cite{APSIPAlyonsrajcheney, Asilomarlyonsrajcheney, CGNetTSP} take $g$ as a steepest descent step and \cite{DRCGNetTCI} uses a ISTA step.

\subsection{Unrolled Deep Neural Network (G-CG-Net)}\label{sec:G-CG-Net}

\begin{figure*}[!t]
    \centering
    \begin{subfigure}{\textwidth}
    \centering
    \includegraphics[scale = 0.65]{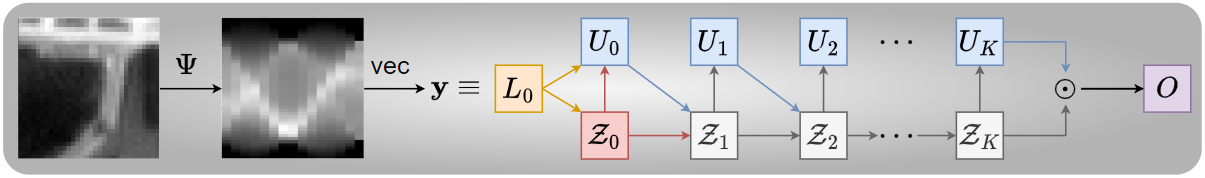}
    \caption{End-to-end network structure of G-CG-Net.}
    \label{fig:DR-CG-Net}
\end{subfigure}
\begin{subfigure}{0.38\textwidth}
    \centering
    \includegraphics[scale = 0.65]{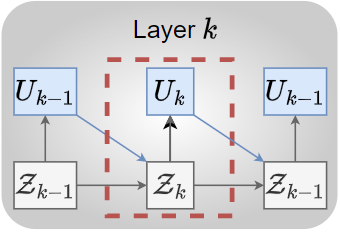}
    \caption{Layer $k$ analogous to iteration $k$ in Algorithm \ref{alg:CG-LS}.}
    \label{fig:DR-CG-Net block}
\end{subfigure}
\begin{subfigure}{0.61\textwidth}
    \centering
    \includegraphics[scale = 0.65]{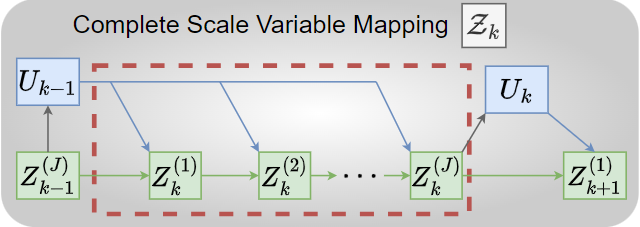}
    \caption{Complete scale variable mapping, $\mathcal{Z}_k$, producing estimate $\bm{z}_k^{(J)}$ in Algorithm \ref{alg:CG-LS}.}
    \label{fig:DR-CG-Net scale mapping module}
\end{subfigure}
\caption{End-to-end network structure for G-CG-Net, the unrolled deep neural network of Algorithm \ref{alg:CG-LS}, is shown in (\ref{fig:DR-CG-Net}). G-CG-Net consists of an input block, $L_0$, initialization block, $\mathcal{Z}_0$, $K+1$ Tikhonov blocks, $U_k$, output block, $O$, and $K$ complete scale variable mappings, $\mathcal{Z}_k$, with structure in (\ref{fig:DR-CG-Net scale mapping module}). Each $\mathcal{Z}_k$ consists of $J$ scale variable updates $Z_k^{(j)}$.}
\label{fig:DR-CG-Net structure}
\end{figure*}

Applying algorithm unrolling to Algorithm \ref{alg:CG-LS}, we create G-CG-Net with end-to-end structure shown in Fig. \ref{fig:DR-CG-Net structure}. Each layer $k$ of G-CG-Net, shown in the dashed box of Fig. \ref{fig:DR-CG-Net block}, corresponds to iteration $k$ of Algorithm \ref{alg:CG-LS} and implements a complete scale variable mapping, $\mathcal{Z}_k$, shown in Fig. \ref{fig:DR-CG-Net scale mapping module}, that updates $\bm{z}$ and a Tikhonov update of $\bm{u}$. Each $\mathcal{Z}_k$ consists of $J$ scale variable updates $Z_k^{(1)}, \ldots, Z_k^{(J)}$ where every $Z_k^{(j)}$ updates $\bm{z}$ once and is the output of a scale-variable-descent update, denoted $g_k^{(j)}$, as given in line \ref{line:z update} of Algorithm \ref{alg:CG-LS}. 

Mathematically detailing the G-CG-Net blocks we have:

\noindent\begin{tikzpicture}
\node[scale = .9, fill = peachfill, draw = darkorangeborder, thick] at (0, 0) {$\phantom{L}$};
\node[scale = 1] at (0, 0) {$L_0$};
\node[anchor = west] at (0.15,0) {$= \bm{y}$ is the input measurements to the network};
\end{tikzpicture} 

\vspace*{-.05cm}

\noindent\begin{tikzpicture}
\node[scale = .9, fill = pinkfill, draw = darkredborder, thick] at (0, 0) {$\phantom{Z}$};
\node[scale = 1] at (0, 0) {$\mathcal{Z}_0$};
\node[anchor = west] at (0.15,0) {$=\mathcal{P}_{0,b}(\hat{A}^T\bm{y})$, for $\hat{A} = \frac{A}{\norm{A}_{2}}$, is an initial estimate of $\bm{z}$.};
\end{tikzpicture} 

\vspace*{-.05cm}

\noindent\begin{tikzpicture}
\node[scale = .9, fill = bluefill, draw = blueborder, thick] at (0, 0) {$\phantom{U}$};
\node[scale = 1] at (0, 0) {$U_k$};
\node[anchor = west] at (0.15,0) {$ = \mathcal{T}_{\bm{y}}(Z_k^{(J)})$ is the Tikhonov estimate of $\bm{u}$};
\node[anchor = west] at (0.15,-.45) {corresponding to line \ref{line:initial} and \ref{line:u update} of Algorithm \ref{alg:CG-LS}.};
\end{tikzpicture}

\vspace*{-.15cm}

\noindent The $k$th complete scale variable mapping, $\mathcal{Z}_k$, contains:

\hspace*{-.16cm}\begin{tikzpicture}
\node[scale = 1.2, fill = greenfill, draw = greenborder, thick] at (0, 0) {$\phantom{Z}$};
\node[scale = 1] at (0, 0) {$Z_k^{(j)}$};
\node[anchor = west] at (0.2,0) {$= \mathcal{P}_{0,z_\infty}(g_k^{(j)}(Z_k^{(j-1)},U_{k-1}))$ is the scale variable};
\node[anchor = west] at (0.2,-.45) {update corresponding to line \ref{line:z update} in Algorithm \ref{alg:CG-LS}.};
\end{tikzpicture}


\noindent\begin{tikzpicture}
\node[scale = .9, fill = purplefill, draw = purpleborder, thick] at (0, 0) {$\phantom{O}$};
\node[scale = 1] at (0, 0) {$O$};
\node[anchor = west] at (0.15,0) {$= \rho_{c_{\max}}(\mathcal{Z}_K\odot U_K)$ is the estimated signal output.};
\end{tikzpicture}

Note that $\rho_{c_{\max}}$ and $\mathcal{P}_{0,z_\infty}$, for $c_{\max} \geq 0$ and $z_{\infty}\geq 0$, are applied for technical reasons discussed in Section \ref{sec:data bounds}. Furthermore, to simplify notation, we let $Z_{k+1}^{(0)} = Z_{k}^{(J)}$ for $k\in\{1, 2, \ldots, K-1\}$ and $Z_1^{(0)} = \mathcal{Z}_0$.

Every $U_k$ layer is parameterized by the same covariance matrix $P_u$ and each $Z_k^{(j)}$ we assume, generally, to be parameterized by $D$ weights $\bm{\theta}_{k,1}^{(j)}, \ldots, \bm{\theta}_{k,D}^{(j)}$. We additionally assume that $\bm{\theta}_{k,d}^{(j)} \in \Omega_d$ for $k\in \mathbb{N}[K]$, $j\in\mathbb{N}[J]$, and some finite-dimensional vector space $\Omega_d$. For instance, if a fully-connected layer, mapping from $\mathbb{R}^n\to\mathbb{R}^n$, is implemented in each $g_k^{(j)}$, then every $\bm{\theta}_{k,1}^{(j)}$ could be the weight matrix from the layer while $\bm{\theta}_{k,2}^{(j)}$ could be the additive bias of the layer. Hence, the G-CG-Net parameters are
\begin{align}
    \bm{\Theta} = \{P_u\}\cup \{\bm{\theta}_{k,d}^{(j)}\}^{j\in\mathbb{N}[J]}_{k\in\mathbb{N}[K],\,\, d\in\mathbb{N}[D].} \label{eqn:G-CG-Net params}
\end{align}

We remark that a structured form can be imposed on $P_u$. In particular, to ensure that $P_u$ is symmetric and positive definite (SPD), we consider, for $\epsilon > 0$ a small fixed real number,
\begin{align*}
    P_u = \begin{cases}
        \max\{\lambda, \epsilon\} I & \textnormal{ Scaled Identity} \\
        \textnormal{diag}([\max\{\lambda_i, \epsilon\}]_{i =1}^n) & \textnormal{ Diagonal} \\
        L_{\textnormal{tri}}L_{\textnormal{tri}}^T + \epsilon I & \textnormal{ Tridiagonal} \\
        LL^T + \epsilon I & \textnormal{ Full}.
    \end{cases}
\end{align*}
In the scaled identity case, only a constant $\lambda$ is learned. In the diagonal case, a vector $\bm{\lambda} = [\lambda_i]_{i = 1}^n$ is learned. In the tridiagonal case, two vectors $\bm{\lambda}_1\in\mathbb{R}^n$ and $\bm{\lambda}_2\in\mathbb{R}^{n-1}$ are learned such that the lower triangular matrix component $L_{\textnormal{tri}}$ is formed by placing $\bm{\lambda}_1$ on the diagonal and $\bm{\lambda}_2$ on the first subdiagonal. Finally, in the case of a full covariance matrix, an entire lower triangular matrix $L$ is learned.

\subsection{Realizations}

Two specific forms of G-CG-Net are detailed here.

\subsubsection{Compound Gaussian Network (CG-Net)~\cite{APSIPAlyonsrajcheney}}

For this method, the scale variable is formulated as $\bm{z} = h(\bm{\rchi})$ for $\bm{\rchi}\sim\mathcal{N}(\bm{0},I)$ and $h$ is a componentwise, non-linear, twice continuously differentiable, and invertible function. Accordingly, the scale variable regularization is $\mathcal{R}(\bm{z}) = \mu\norm{h^{-1}(\bm{z})}_2^2$, for a scalar constant $\mu > 0$, to enforce normality of $\bm{\rchi} = h^{-1}(\bm{z})$. 

The scale-variable-descent update, $g$, is a projected steepest descent step based on a learned quadratic norm. We consider a slightly adjusted update, to assist in the analysis, given as
\begin{align}
   g_k^{(j)}(\bm{z},\bm{u}) = \mathcal{P}_{a,b}(\bm{z} - B_{k}^{(j)} \rho_{\xi}(\nabla_{\bm{z}} F(\bm{u},\bm{z}; \mu_k^{(j)}))) \label{eqn:CG-Net scale variable descent update}
\end{align}
where $a, b, \xi > 0$ are fixed real-valued scalars, $B_k^{(j)}$ is a learned $n\times n$ positive definite matrix, and 
\begin{align*}
   \nabla_{\bm{z}} F(\bm{u},\bm{z}; \mu_k^{(j)}) = A_{\bm{u}}^T(A_{\bm{u}}\bm{z}-\bm{y}) + \mu_k^{(j)} [h^{-1}]'(\bm{z})\odot h^{-1}(\bm{z})
\end{align*}
for $\mu_k^{(j)}$ a learned scalar. Note, the application of $\rho_{\xi}$ ensures a sufficiently small step size is used in each gradient update for numerical stability.

For parameters, $P_u$ is structured as a scaled identity matrix and $D = 2$ where $\bm{\theta}_{k,1}^{(j)} = B_k^{(j)}$ and $\bm{\theta}_{k,2}^{(j)} = \mu_k^{(j)}$. A structural similarity index measure (SSIM) loss function is used to train CG-Net. For two images or matrices $I_1$ and $I_2$ of equivalent size, the SSIM loss function is given by $L(I_1, I_2) = 1 - \textnormal{SSIM}(I_1, I_2)$.

\subsubsection{Deep Regularized Compound Gaussian Network (DR-CG-Net)~\cite{Asilomarlyonsrajcheney}} In this method, the scale variable regularization is left as an implicit function that is learned, through its gradient, in the unrolled deep neural network. In DR-CG-Net, a projected gradient descent (PGD) or ISTA step are employed as scale-variable-descent updates. Due to similar analyses of both scale-variable-descent updates, we focus on the PGD scale update method given by
\begin{align}
    g_k^{(j)}(\bm{z},\bm{u}) =
   v_k^{(j)}(\bm{z},\bm{u};\delta_k^{(j)}) + \mathcal{V}_k^{(j)}\left(\bm{z}\right)  \label{eqn:PGD scale variable update}
\end{align}
where, for a step size $\delta_k^{(j)}> 0$ and fixed real number $\xi > 0$,
\begin{align}
    v_k^{(j)}(\bm{z},\bm{u}; \delta_k^{(j)}) = \bm{z} - \delta_k^{(j)}\rho_{\xi}(A_{\bm{u}}^T(A_{\bm{u}}\bm{z}-\bm{y})) \label{eqn:data fidelity gradient update}
\end{align}
is a gradient update of $\bm{z}$ over the data fidelity term of (\ref{eqn:cost function}) and $\mathcal{V}_k^{(j)}:\mathbb{R}^n\to\mathbb{R}^n$ an embedded subnetwork. We remark that the update in (\ref{eqn:PGD scale variable update}) is a gradient descent step on (\ref{eqn:cost function}) w.r.t.~$\bm{z}$  when $\mathcal{V}_k^{(j)} = \nabla \mathcal{R}$. Hence, in training $\mathcal{V}_k^{(j)}$, the regularization, or equivalently prior distribution, for the scale variable is learned. Finally, $\rho_{\xi}$ is applied for numerical stability as in CG-Net.

Each subnetwork, $\mathcal{V}_k^{(j)}$, consists of $L_c$ convolutional layers using ReLU activation functions.  That is, layer $\ell$ consists of $f_\ell$ convolutions, i.e. filter channels, using kernel size $k_\ell\times k_\ell$ with unit stride. Note, zero padding is applied to each filter channel of the input such that the output, at any filter channel, is the same size as the input to any filter channel. Furthermore, we take $f_{L_c} = 1$ so that given a single channel image input to $\mathcal{V}_k^{(j)}$ the output is also a single channel image of equivalent dimension. For our analysis, we assume, without loss of generality, that convolutional layer $\ell$ of $\mathcal{V}_k^{(j)}$ is implemented as a matrix-vector product with weight matrix $W_{k,\ell}^{(j)}$. 

For parameters, $P_u$ is structured as a tridiagonal matrix and $D = L_c+1$ where $\bm{\theta}_{k, \ell}^{(j)} = W_{k, \ell}^{(j)}$ for $1\leq \ell\leq L_c$ and $\bm{\theta}_{k, L_c+1}^{(j)} = \delta_k^{(j)}$. A mean-absolute error loss function, $L(\bm{x}_1,\bm{x}_2) = \frac{1}{n}\norm{\bm{x}_1 - \bm{x}_2}_1$, is used to train DR-CG-Net.

\section{Generalization Error Bounds}

In order to estimate the generalization error we, similarly to~\cite{NNfromstatistical,kouni2022deconet,joukovsky2021generalization}, derive an upper bound on $\mathcal{L}(\widehat{\bm{c}})$ in terms of $\mathcal{L}_{\mathcal{S}}(\widehat{\bm{c}})$ and a Dudley's inequality bound of the Rademacher complexity for the hypothesis space generated by G-CG-Net. The Dudley's inequality bound is evaluated using a covering number argument dependent on a Lipschitz property for G-CG-Net outputs w.r.t. G-CG-Net parameters. Our key contribution here is showing the G-CG-Net outputs are indeed Lipschitz w.r.t. to the network parameters and applying our derived GEB to the specific CG-Net and DR-CG-Net structures. 

The remainder of this section is structured as follows. Section \ref{sec:data bounds} explicates boundedness assumptions that underlie our GEBs for deep compound Gaussian networks. We detail a GEB for G-CG-Net in section \ref{sec:GEB G-CG-Net}, which is subsequently refined for the CG-Net and DR-CG-Net structures in sections \ref{sec:GEB CG-Net} and \ref{sec:GEB DR-CG-Net}, respectively.

\subsection{Boundedness Assumptions} \label{sec:data bounds}

A common assumption in machine learning literature and implementation is that the input data to a DNN is bounded. Specific bounds are often guaranteed through preprocessing of the data, which has been shown to assist in the performance of DNN models~\cite{sola1997datanormalizationimportance}. For instance, in~\cite{APSIPAlyonsrajcheney, Asilomarlyonsrajcheney}, which use images as the signals of interest, each image is scaled down to be bounded in the Euclidean unit ball. Furthermore, bounded data implies that the possible parameters to be learned by a DNN are similarly bounded as DNNs are trained only for a finite number of epochs using a small learning rate to estimate bounded signals from bounded inputs. 
\begin{assumption} \label{assumption:data bound}
    The following bounds hold almost surely:
    \begin{enumerate}
        \item Original signals, $\bm{c}\in\mathbb{R}^n$, satisfy $\norm{\bm{c}}_2\leq c_{\max}$.
        \item Scale variables, $\bm{z}\in\mathbb{R}^n$, satisfy $\norm{\bm{z}}_{\infty} \leq z_{\infty}$.
        \item Covariance matrices, $P_u$, satisfy $P_u\in \mathcal{P}$ where, for bounding scalars $0 < p_{\min} \leq p_{\max}$, 
        \end{enumerate}
       \begin{align*}
            \mathcal{P} &= \{n\times n \textnormal{ SPD matrices with bounded spectrum}\} \\
              &= \{P\in\mathbb{R}^{n\times n}:\textnormal{ SPD, } \norm{P}_{2} \leq p_{\max}, \norm{P^{-1}}_{2} \leq p_{\min}^{-1}\}.
        \end{align*}
        \begin{enumerate}
            \item[4)] Each scale variable update parameter, $\bm{\theta}_{k, d}^{(j)}$, satisfies $\norm{\bm{\theta}_{k, d}^{(j)}}_{(d)} \leq \omega_{d}$ for scalar $\omega_{d} \geq 0$ and some norm $\norm{\cdot}_{(d)}.$
        \end{enumerate}
\end{assumption}

Note that we assume the scale variables are bounded since the original signals are bounded. This is enforced by the mReLU activation function, $\mathcal{P}_{0,z_{\infty}}$, in each scale variable update layer $Z_k^{(j)}$. Finally, as the original signals are bounded, we force the G-CG-Net estimates to be equivalently bounded by applying the projection operator, $\rho_{c_{\max}}$, to $\mathcal{Z}_K\odot U_K$.

\subsection{G-CG-Net} \label{sec:GEB G-CG-Net}

To derive a GEB for G-CG-Net, we require the an assumption on the loss function and scale-variable-descent update.

\begin{assumption} \label{assumption:loss function}
For a vector space $V$, the loss function $L(\bm{x}_1, \bm{x}_2):V\to\mathbb{R}$ satisfies:
\begin{align*}
    &1. & &\textnormal{Bounded:} & &|L(\bm{x}_1, \bm{x}_2)|\leq c \\
    &2. & &\tau\textnormal{-Lipschitz:} & &\norm{L(\bm{x}_1, \bm{x}) - L(\bm{x}_2, \bm{x})}_2 \leq \tau \norm{\bm{x}_1-\bm{x}_2}_2
\end{align*}
for $c\geq 0$, $\tau\geq 0$ and all $\bm{x}_1, \bm{x}_2$, $\bm{x}\in V$.
\end{assumption}

\begin{assumption}\label{assumption:scale update method}
    For every $\bm{z}_i\in\mathbb{R}^n$ satisfying $\norm{\bm{z}_i}_\infty \leq z_\infty$ and $\bm{u}_i = \mathcal{T}_{\bm{y}}(\bm{z}_i; P_i)$ where $P_i\in\mathcal{P}$, each $g_k^{(j)}(\bm{z},\bm{u})$, parameterized by $\vartheta_k^{(j)} = (\bm{\theta}_{k,1}^{(j)}, \ldots, \bm{\theta}_{k,D}^{(j)})$, satisfies
        \begin{multline*}
        \norm{g_k^{(j)}(\bm{z}_1,\bm{u}_1; \vartheta_k^{(j)}) - g_k^{(j)}(\bm{z}_2,\bm{u}_2; \widetilde{\vartheta}_{k}^{(j)})}_2 \\
        \leq r_{k,1}^{(j-1)} \norm{\bm{z}_1 - \bm{z}_2}_2 + r_{k,2}^{(j-1)} \norm{\bm{u}_1 - \bm{u}_2}_2  \\
        +\sum_{d = 1}^D r_{k,d,3}^{(j-1)} \norm{\bm{\theta}_{k,d}^{(j)} - \widetilde{\bm{\theta}}_{k,d}^{(j)}}_{(d)}
    \end{multline*}
    for non-negative scalar constants $r_{k,1}^{(j-1)}, r_{k,2}^{(j-1)}$, $r_{k,d,3}^{(j-1)}$, and norms $\{\norm{\cdot}_{(d)}\}_{d = 1}^D$.
\end{assumption}

While the Lipschitz property of Assumption \ref{assumption:scale update method} may seem arbitrarily restrictive, we show it holds for CG-Net and DR-CG-Net in Appendix \ref{apndx:CG-Net GEB proof} and Appendix \ref{apndx:DR-CG-Net GEB proof}, respectively.

Next, define $\mathcal{P}_{\textnormal{const}}\subset \mathcal{P}_{\textnormal{diag}}\subset \mathcal{P}_{\textnormal{tri}} \subset \mathcal{P}_{\textnormal{full}}$ to be the options for $\mathcal{P}$ corresponding to the vector spaces of constant, diagonal, tridiagonal, and full covariance matrices with bounded spectrum, respectively. Additionally, let each scale variable update parameter, $\bm{\theta}_{k,d}^{(j)}$, be of dimension $\alpha_{d} \geq 0$ (i.e. $\bm{\theta}_{k,d}^{(j)}\in\mathbb{R}^{\alpha_{d}}$) and define the sets
\begin{align}
    \Omega_{d} = \{\bm{\theta} \in\mathbb{R}^{\alpha_{d}}: \norm{\bm{\theta}}_{(d)} \leq \omega_{d}\}. \label{eqn:scale variable update parameter spaces}
\end{align}
Then the hypothesis class for G-CG-Net is
\begin{align*}
    \resizebox{\columnwidth}{!}{${\displaystyle\mathcal{H}_{\textnormal{CG}}^{(1)} = \left\{\widehat{\bm{c}}\left(\boldsymbol{\cdot}; \{P_u,\bm{\theta}_{k,d}^{(j)}\}\middlescript{\substack{j \in\mathbb{N}[J] \\ k \in\mathbb{N}[K] \\ d \in\mathbb{N}[D]}}\right): P_u \in\mathcal{P}, \bm{\theta}_{k,d}^{(j)}\in\Omega_{d}\right\}}$.}
\end{align*}

\begin{theorem}[Generalization Error Bound for G-CG-Net] \label{thm:GEB G-CG-Net}
Let $\mathcal{S} = \{(\overline{\bm{y}}_i, \overline{\bm{c}}_i)\}_{i = 1}^{N_s}$ be a training dataset where each $(\overline{\bm{c}}_i, \overline{\bm{y}}_i)$ is given by (\ref{eqn:linear_msrmt}) and define $y_{\max} = \max_{1\leq i\leq N_s} \norm{\overline{\bm{y}}_i}_2.$ If Assumption \ref{assumption:data bound}, \ref{assumption:loss function}, and \ref{assumption:scale update method} hold then with probability at least $1-\varepsilon$, for all $\widehat{\bm{c}}\in\mathcal{H}_{\textnormal{CG}}^{(1)}$, the generalization error of G-CG-Net is bounded as
             \begin{align*}
             &\mathcal{L}(\widehat{\bm{c}})\leq \mathcal{L}_{\mathcal{S}}(\widehat{\bm{c}}) + \\
             &  \frac{8\tau c_{\max}}{\sqrt{N_s}}\left(\sqrt{\dim(\mathcal{P})}\sqrt{\ln\left(e\left(1+\frac{4p_{\max}(KJD+1)\kappa}{c_{\max}}\right)\right)}\right. \\
             & \resizebox{\columnwidth}{!}{${\displaystyle +\left.\sum_{k = 1}^K\sum_{j = 1}^J\sum_{d = 1}^D\sqrt{\alpha_{d}\ln\left(e\left(1+\frac{4\omega_{d}(KJD+1)\kappa_{k,d}^{(j)}}{c_{\max}}\right)\right)}\right)}$} \\
             & + 4c\sqrt{2\ln(4/\varepsilon)/N_s}
        \end{align*}
   for $\dim(\mathcal{P}) = 1, n, 2n-1,$ or $n(n+1)/2$ when $\mathcal{P} = \mathcal{P}_{\textnormal{const}}$, $\mathcal{P} = \mathcal{P}_{\textnormal{diag}}$, $\mathcal{P} = \mathcal{P}_{\textnormal{tri}}$, or $\mathcal{P} = \mathcal{P}_{\textnormal{full}}$, respectively. Additionally, 
    \begin{align}
        \kappa &= z_{\infty}(c_{1}+p_{\max}y_{\max} \norm{A}_{\infty})\widehat{c}_{1}^{(K,J)} + z_{\infty} c_{2} \label{eqn:kappa}\\
        \kappa_{k,d}^{(j)} &=  z_{\infty}(c_{1}+p_{\max}y_{\max} \norm{A}_{\infty})\widehat{c}_{k,j,d,2}^{(K,J)} \label{eqn:kappa k,d,j}
    \end{align}
    where
    \begin{align*}
        c_{1} &= p_{\max}y_{\max}\,\, \norm{A}_2\left(1 + 2z_{\infty}^2p_{\max}\,\, \norm{A}_2^2\right) \\
    c_{2} &= z_{\infty}y_{\max} \norm{A}_2\left(p_{\max}/p_{\min}\right)^2 \\
    \widehat{c}_{1}^{(K,J)} &= c_{2}\sum_{k = 1}^K \widehat{r}_{k,2}^{(J)} \prod_{\ell = k+1}^K (\widehat{r}_{\ell,1}^{(J)}+\widehat{r}_{\ell,2}^{(J)} c_{1}) \\
    \widehat{c}_{k,j,d,2}^{(K,J)} &= \widehat{r}_{k,d,3}^{(j, J)}\prod_{\ell = k+1}^K (\widehat{r}_{\ell,1}^{(J)}+\widehat{r}_{\ell,2}^{(J)} c_{1})
    \end{align*}
    for ${\displaystyle \hspace{.25cm} \widehat{r}_{k,1}^{(J)} = \prod_{j = 1}^{J} r_{k,1}^{(j-1)}, \hspace{.5cm} \widehat{r}_{k,2}^{(J)} = \sum_{j = 1}^{J} r_{k,2}^{(j-1)}\prod_{\ell = j}^{J-1} r_{k,1}^{(\ell)}}$, \hspace*{\fill} and ${\displaystyle \widehat{r}_{k,d,3}^{(j,J)} = r_{k,d,3}^{(j-1)}\prod_{\ell = j}^{J-1} r_{k,1}^{(\ell)}}.$
\end{theorem}

A proof of Theorem \ref{thm:GEB G-CG-Net} is provided in Appendix \ref{apndx:G-CG-Net GEB proof}, which broadly employs Dudley's inequality to bound the Rademacher complexity of $\mathcal{H}_{\textnormal{CG}}^{(1)}$. Ideally, through training G-CG-Net, a hypothesis $\widehat{\bm{c}}\in\mathcal{H}_{\textnormal{CG}}^{(1)}$ is chosen such that $\mathcal{L}_{\mathcal{S}}(\widehat{\bm{c}})$ is minimized, but any $\widehat{\bm{c}}$ possibly generated by G-CG-Net could be used in the Theorem \ref{thm:GEB G-CG-Net} bound. For instance, if early stopping is implemented in training then Theorem \ref{thm:GEB G-CG-Net} still applies to the generated $\widehat{\bm{c}}$ despite $\widehat{\bm{c}}$ not optimizing $\mathcal{L}_{\mathcal{S}}$.

Lastly, we remark for noiseless measurements, that is $\overline{\bm{y}}_i = A\overline{\bm{c}}_i$, that $y_{\max} \leq c_{\max} \norm{A}_2$ for any set of training data. For white noise measurements, that is $\overline{\bm{y}}_i = A\overline{\bm{c}}_i + \bm{\nu}$ where $\bm{\nu}\sim \mathcal{N}(0,\sigma^2 I)$, then $y_{\max} \leq c_{\max} \norm{A}_2 + z_p\sigma$ with high probability for large $z_p$. For instance, using the quantile function of a normal distribution, $y_{\max} \leq c_{\max} \norm{A}_2 + 6.11\sigma$ with probability $(1-2\times 10^{-9})^{mN_s} \approx 1-2mN_s\times 10^{-9}$. 

\subsection{CG-Net} \label{sec:GEB CG-Net}

For CG-Net, the scale variable update parameter spaces, $\Omega_{d}$ from (\ref{eqn:scale variable update parameter spaces}), are $\Omega_{1} = \mathcal{P}_{\textnormal{full}}$ and $\Omega_{2} = [-\mu,\mu]$ for constant $\mu > 0$. Thus, the hypothesis class for CG-Net is
\begin{align*}
    \mathcal{H}_{\textnormal{CG}}^{(2)} = \left\{\widehat{\bm{c}}\left(\boldsymbol{\cdot}; \left\{P_u,B_k^{(j)}, \mu_k^{(j)}\right\}_{\substack{k \in\mathbb{N}[K]}}^{j \in\mathbb{N}[J]}\right): P_u\in\mathcal{P}_{\textnormal{const}},\right. &\\
  \left.  B_k^{(j)}\in\Omega_{1}, \mu_k^{(j)}\in \Omega_{2}\right\}&.
\end{align*}

\begin{theorem}[Generalization Error Bound for CG-Net]
Let $\mathcal{S} = \{(\overline{\bm{y}}_i, \overline{\bm{c}}_i)\}_{i = 1}^{N_s}$ be a training dataset where each $(\overline{\bm{c}}_i, \overline{\bm{y}}_i)$ is given by (\ref{eqn:linear_msrmt}) and define $y_{\max} = \max_{1\leq i\leq N_s} \norm{\overline{\bm{y}}_i}_2.$ If Assumption \ref{assumption:data bound} holds then with probability at least $1-\varepsilon$, for all $\widehat{\bm{c}}\in\mathcal{H}_{\textnormal{CG}}^{(2)}$, the generalization error of CG-Net is bounded as in Theorem \ref{thm:GEB G-CG-Net} for $\tau$ a Lipschitz constant of the SSIM loss function, $\dim(\mathcal{P}) = 1$, $c = 2$, $D = 2$,
\begin{align*}
    (\alpha_{d}, \omega_{d}) = \begin{cases}
        (\frac{n(n+1)}{2}, p_{\max}) & d = 1 \\
        (1, \mu) & d = 2,
    \end{cases}
\end{align*}
${\displaystyle \widehat{r}_{k,1}^{(J)} = r_1^{J}, \hspace{.5cm} \widehat{r}_{k,2}^{(J)} = r_2 \frac{1-r_1^J}{1-r_1}},$ and
\begin{align}
        \widehat{r}_{k,d,3}^{(j,J)} = r_1^{J-j}\begin{cases}
            \xi  & d = 1 \\
            p_{\max}h_{\max} & d = 2
        \end{cases} \label{eqn:r_3 CG-Net}
    \end{align}
where
\begin{align*}
    &r_1 = 1+p_{\max}((z_{\infty} p_{\max} y_{\max}\norm{A}_2 \norm{A}_{\infty})^2 + \mu\tau_h) \\
    &\resizebox{\columnwidth}{!}{${\displaystyle r_2 = p_{\max} y_{\max} \norm{A}_2\left(1+z_{\infty}^2p_{\max} \norm{A}_2 (\norm{A}_2+\norm{A}_{\infty})\right) }$} \\
    &h_{\max} = \max_{z\in [a,b]}\,\, [h^{-1}]'(z)h^{-1}(z),
\end{align*}
and $\tau_h = \max_{z\in [a,b]}\,\, [h^{-1}]''(z)h^{-1}(z) + [h^{-1}]'(z)^2.$
\label{thm:GEB CG-Net}
\end{theorem}

We remark that in Theorem \ref{thm:GEB CG-Net} we only state a Lipschitz constant for the SSIM loss function exists as deriving one is space consuming and not illuminating. This is, in part, due to the fact that SSIM$(I_1,I_2)$, from the SSIM loss function, is implemented as the average structural similarity over a set of patches from the input images where in each patch a Gaussian weighting filter is used~\cite{NN_loss}.

In all numerical experiments in~\cite{APSIPAlyonsrajcheney, Asilomarlyonsrajcheney}, $h(z) = \exp(z)$ on $[a,b] = [1,\exp(3)]$ and $\xi = 1$. Thus, $h_{\max} = \exp(-1)$, $\tau_h = 1$, and $z_{\infty} = \exp(3)$. Furthermore, preprocessing is used such that $c_{\max} = 1$ and for $\epsilon > 0$, a small stabilizing parameter, $p_{\max} = 1/\epsilon$, $p_{\min} = \epsilon$, and $|\mu| \leq 1/\epsilon$. The remaining constants $\norm{A}_2$, $\norm{A}_{\infty}$, and $y_{\max}$ can be calculated given the measurement model and training dataset.

Let $a\lesssim b$ imply $a\leq s_cb$ for some $s_c > 0$ and define
\begin{align}
    r = \ln(y_{\max})+\ln(\norm{A}_2)+\ln(\norm{A}_{\infty}). \label{eqn:r}
\end{align}
\begin{corollary}\label{corollary:big O GEB CG-Net}
    The generalization error for CG-Net scales as
    \begin{align*}
       |\mathcal{L}(\widehat{\bm{c}}) - \mathcal{L}_{\mathcal{S}}(\widehat{\bm{c}})| \lesssim n\sqrt{\frac{(KJ)^3r}{N_s}}. 
    \end{align*}
    Furthermore, as, $y_{\max} \lesssim \sqrt{m}$, $\norm{A}_2\lesssim n\sqrt{m}$, and $\norm{A}_{\infty}\lesssim n$ then the CG-Net generalization error scales at most as
    \begin{align*}
        |\mathcal{L}(\widehat{\bm{c}}) - \mathcal{L}_{\mathcal{S}}(\widehat{\bm{c}})| \lesssim n\sqrt{\frac{(KJ)^3(\ln(m)+\ln(n))}{N_s}}.
    \end{align*}
\end{corollary}
Corollary \ref{corollary:big O GEB CG-Net} results by ignoring constants in the GEB from Theorem \ref{thm:GEB CG-Net} to consider how this GEB scales in network size and signal dimension. From Corollary \ref{corollary:big O GEB CG-Net}, once the amount of training data satisfies $N_s\sim n^2(KJ)^3(\ln(m)+\ln(n))$ then the GEB of CG-Net will be small with high probability.

\subsection{DR-CG-Net} \label{sec:GEB DR-CG-Net}

For DR-CG-Net, the scale variable parameter spaces are
\begin{align*}
    \Omega_{d} = \begin{cases}
        \{W\in\mathbb{R}^{nf_{d}\times nf_{d-1}}: \norm{W}_{2} \leq w_d\} & d\in \mathbb{N}[L_c] \\
        [-\delta, \delta] & d = L_c+1.
    \end{cases}
\end{align*}
for real value constants $w_d, \delta > 0.$
Note, each $W\in\Omega_{d}$, for $d\in\mathbb{N}[L_c]$, corresponds to a convolutional layer mapping from $f_{d-1}$ to $f_d$ filter channels using convolutional kernels of size $k_d\times k_d$. Thus, the hypothesis class for DR-CG-Net is
\begin{align*}
    \mathcal{H}_{\textnormal{CG}}^{(3)} = \left\{\bm{c}\left(\boldsymbol{\cdot}; \{P_u,\delta_k^{(j)}, W_{k,\ell}^{(j)}\}\middlescript{\substack{j \in\mathbb{N}[J] \\ k \in\mathbb{N}[K] \\ \ell \in\mathbb{N}[L_c]}}\right): P_u\in\mathcal{P}_{\textnormal{tri}},\right. & \\
    \hspace{.25cm}\left. W_{k,\ell}^{(j)}\in \Omega_{\ell},  \delta_k^{(j)}\in \Omega_{L_c+1}\right\}.&
\end{align*}

\begin{theorem}[Generalization Error Bound for DR-CG-Net]
    Let $\mathcal{S} = \{(\overline{\bm{y}}_i, \overline{\bm{c}}_i)\}_{i = 1}^{N_s}$ be a training dataset where each $(\overline{\bm{c}}_i, \overline{\bm{y}}_i)$ is given by (\ref{eqn:linear_msrmt}) and define $y_{\max} = \max_{1\leq i\leq N_s} \norm{\overline{\bm{y}}_i}_2.$ If Assumption \ref{assumption:data bound} holds then with probability at least $1-\varepsilon$, for all $\widehat{\bm{c}}\in\mathcal{H}_{\textnormal{CG}}^{(3)}$, the generalization error of DR-CG-Net is bounded as in Theorem \ref{thm:GEB G-CG-Net} with $\tau = 1/\sqrt{n}$, $\dim(\mathcal{P}) = 2n-1$, $c = c_{\max}/\sqrt{n}$, $D = L_c+1$,
    \begin{align}
       (\alpha_{d}, \omega_{d}) &= \begin{cases}
            (f_{d-1}f_dk_d^2, w_d) & d = 1, 2, \ldots, L_c \\
            (1, \delta) & d = L_c+1,
        \end{cases} \label{eqn:alpha, omega}
    \end{align}
    ${\displaystyle \widehat{r}_{k,1}^{(J)} = r_1^{J}, \hspace{.5cm} \widehat{r}_{k,2}^{(J)} = r_2 \frac{1-r_1^J}{1-r_1}},$ and
    \begin{align}
        \widehat{r}_{k,d,3}^{(j,J)} = r_1^{J-j}\begin{cases}
            \left(\sqrt{n} z_{\infty}\prod_{\substack{\ell = 1 \\ \ell\neq d}}^{L_c}w_{\ell}\right) & d = 1, 2, \ldots, L_c \\
            \xi  & d = L_c+1
        \end{cases} \label{eqn:r_3 DR-CG-Net}
    \end{align}
where
\begin{align*}
    r_1 &= 1+\delta (z_{\infty} p_{\max} y_{\max} \norm{A}_2 \norm{A}_{\infty})^2 + \prod_{\ell = 1}^{L_c} w_{\ell} \\
     r_2 &= \delta y_{\max} \norm{A}_2\left(1+z_{\infty}^2p_{\max} \norm{A}_2 (\norm{A}_2+\norm{A}_{\infty})\right).
\end{align*}
\label{thm:GEB DR-CG-Net}
\end{theorem}

Ignoring constants, the GEB of DR-CG-Net scales as in the following Corollary.
\begin{corollary} \label{corollary:big O GEB DR-CG-Net}
The GEB for DR-CG-Net scales as
\begin{align*}
    &|\mathcal{L}(\widehat{\bm{c}}) - \mathcal{L}_{\mathcal{S}}(\widehat{\bm{c}})| \lesssim KJr_f(L_c)\sqrt{\frac{\ln(n)}{n N_s}} \hspace{.1cm} + \\ 
    & \resizebox{\columnwidth}{!}{${\displaystyle (\sqrt{n} + KJ(1+r_f(L_c)))\sqrt{\frac{\ln(KJL_c) + KJ(r + \ln(1+\prod_{\ell=1}^{L_c}w_\ell))}{n N_s}} }$}
\end{align*}
for $r_f(L_c) =  \sum_{\ell = 1}^{L_c} \sqrt{f_{\ell-1}f_{\ell}k_{\ell}^2}$ and $r$ given in (\ref{eqn:r}). 

Furthermore, bounding each $w_{\ell}$, $f_{\ell}$, and $k_{\ell}$ respectively by the maximums $w_{\max} = \max_{1\leq \ell\leq L_c} w_{\ell}$, $f_{\max} =\max_{1\leq \ell\leq L_c} f_{\ell}$, and $k_{\max} = \max_{1\leq \ell\leq L_c} f_{\ell}$ and noting that $y_{\max} \lesssim \sqrt{m}$, $\norm{A}_2\lesssim n\sqrt{m}$, and $\norm{A}_{\infty}\lesssim n$ then the GEB of DR-CG-Net scales at most as
    \begin{align*}
        |\mathcal{L}(\widehat{\bm{c}}) - \mathcal{L}_{\mathcal{S}}(\widehat{\bm{c}})| \lesssim \sqrt{\frac{(KJL_c)^3(\ln(m)+\ln(n))}{N_s}}.
    \end{align*}
\end{corollary}

\textbf{Observation:} The CG-Net network size is $\approx KJ$ and DR-CG-Net network size is $\approx KJL_c$. Thus, the GEB of CG-Net further simplifies to $|\mathcal{L} - \mathcal{L}_{\mathcal{S}}|\lesssim n\sqrt{\frac{(\textnormal{Network Size})^3(\ln(m)+\ln(n))}{N_s}}$ while the GEB of DR-CG-Net further simplifies to \hspace{1cm} $|\mathcal{L} - \mathcal{L}_{\mathcal{S}}|\lesssim \sqrt{\frac{(\textnormal{Network Size})^3(\ln(m)+\ln(n))}{N_s}}$. Hence, constraining CG-Net and DR-CG-Net to an equivalent network size, the discrepancy in the GEBs is only the signal dimension where CG-Net scales as $\mathcal{O}(n\sqrt{\ln(n)})$ and DR-CG-Net scales as $\mathcal{O}(\sqrt{\ln(n)})$. Therefore, DR-CG-Net produces a tighter GEB, which is supported by the numerical experiments of~\cite{CGNetTSP, DRCGNetTCI} that show DR-CG-Net produces test reconstructions of superior quality as compared to those produced by CG-Net.

\section{Lipschitz Property of G-CG-Net} \label{sec:lipschitz of G-CG-Net}

In this section, we show that G-CG-Net is Lipschitz w.r.t. its parameters $\bm{\Theta}$ in (\ref{eqn:G-CG-Net params}). This result is a cornerstone of proving Theorem \ref{thm:GEB G-CG-Net} and is dependent on a Lipschitz and bounded property of the Tikhonov solution along with Assumption \ref{assumption:scale update method}.

\subsection{Lipschitz of Complete Scale Variable Mappings}

For notation we write $\mathcal{Z}_k^{(J)}$ as the complete scale variable mapping consisting of $J$ scale variable updates. That is $\mathcal{Z}_k^{(J)} = Z_k^{(J)}\circ \cdots \circ Z_k^{(1)}$.
\begin{prop}\label{prop:scale variable lipschitz 1}
    Let Assumption \ref{assumption:scale update method} hold. Then the complete scale variable mapping $\mathcal{Z}_k^{(J)}(\bm{z}, \bm{u})$, which is parameterized by some $\Theta_k^{(J)} = (\bm{\theta}_{k,1}^{(j)}, \ldots, \bm{\theta}_{k,D}^{(j)})_{j \in\mathbb{N}[J]}$, satisfies 
    \begin{align}
        \resizebox{\columnwidth}{!}{${\displaystyle\norm{\mathcal{Z}_k^{(J)}(\bm{z}_1,\bm{u}_1; \Theta_k^{(J)}) - \mathcal{Z}_k^{(J)}(\bm{z}_2,\bm{u}_2; \widetilde{\Theta}_k^{(J)})}_2 \leq \widehat{r}_{k,1}^{(J)} \norm{\bm{z}_1-\bm{z}_2}_2}$}& \nonumber \\
          + \widehat{r}_{k,2}^{(J)} \norm{\bm{u}_1 - \bm{u}_2}_2 + \sum_{j = 1}^J\sum_{d = 1}^D \widehat{r}_{k,d,3}^{(j,J)} \norm{\bm{\theta}_{k, d}^{(j)} - \widetilde{\bm{\theta}}_{k,d}^{(j)}}_{(d)}& \nonumber
    \end{align}
    for ${\displaystyle \hspace{.5cm} \widehat{r}_{k,1}^{(J)} = \prod_{j = 1}^{J} r_{k,1}^{(j-1)}, \hspace{.5cm} \widehat{r}_{k,2}^{(J)} = \sum_{j = 1}^{J} r_{k,2}^{(j-1)}\prod_{\ell = j}^{J-1} r_{k,1}^{(\ell)}}$, and
    ${\displaystyle \widehat{r}_{k,d,3}^{(j,J)} = r_{k,d,3}^{(j-1)}\prod_{\ell = j}^{J-1} r_{k,1}^{(\ell)}}.$
\end{prop}
\begin{proof}
    Using induction on $J$, the base case, $J = 1$, holds by Assumption \ref{assumption:scale update method} where we set $\prod_{j= J+1}^J r_{k,1}^{(j)} = 1$. Let the induction hypothesis hold for fixed $J > 1.$ Using Assumption \ref{assumption:scale update method}, the induction hypothesis, and that $\mathcal{P}_{0,z_{\infty}}$ is 1-Lipschitz
    \begin{align}
        &\norm{\mathcal{Z}_k^{(J+1)}(\bm{z}_1,\bm{u}_1; \Theta_k^{(J+1)}) - \mathcal{Z}_k^{(J+1)}(\bm{z}_2,\bm{u}_2; \widetilde{\Theta}_k^{(J+1)})}_2 \nonumber \\
        &= \left\lVert\mathcal{P}_{0,z_{\infty}}(g_k^{(J+1)}(\mathcal{Z}_k^{(J)}(\bm{z}_1,\bm{u}_1; \Theta_k^{(J)}), \bm{u}_1; \vartheta_k^{(J+1)})) \right. \nonumber \\
        &\hspace{.6cm} \left. - \mathcal{P}_{0,z_{\infty}}(g_k^{(J+1)}(\mathcal{Z}_k^{(J)}(\bm{z}_2,\bm{u}_2; \widetilde{\Theta}_k^{(J)}), \bm{u}_2; \widetilde{\vartheta}_k^{(J+1)}))\right\rVert_2 \nonumber \\
        &\leq r_{k,1}^{(J)} \norm{\mathcal{Z}_k^{(J)}(\bm{z}_1,\bm{u}_1; \Theta_k^{(J)}) - \mathcal{Z}_k^{(J)}(\bm{z}_2,\bm{u}_2; \widetilde{\Theta}_k^{(J)})}_2 \nonumber \\
        &\hspace{.35cm} + r_{k,2}^{(J)} \norm{\bm{u}_1-\bm{u}_2}_2 + \sum_{d = 1}^D r_{k,d,3}^{(J)} \norm{\bm{\theta}_{k,d}^{(J+1)} - \widetilde{\bm{\theta}}_{k,d}^{(J+1)}}_{(d)} \nonumber \\
        &\leq r_{k,1}^{(J)}\left(\widehat{r}_{k,1}^{(J)} \norm{\bm{z}_1-\bm{z}_2}_2 + \widehat{r}_{k,2}^{(J)} \norm{\bm{u}_1 - \bm{u}_2}_2\right) \nonumber \\
        & \hspace{.35cm}+ r_{k,1}^{(J)} \sum_{j = 1}^J\sum_{d = 1}^D \widehat{r}_{k,d,3}^{(j,J)} \norm{\bm{\theta}_{k, d}^{(j)} - \widetilde{\bm{\theta}}_{k,d}^{(j)}}_{(d)} + r_{k,2}^{(J)} \norm{\bm{u}_1-\bm{u}_2}_2 \nonumber \\
        & \hspace{.5cm} + \sum_{d = 1}^D r_{k,d,3}^{(J)} \norm{\bm{\theta}_{k,d}^{(J+1)} - \widetilde{\bm{\theta}}_{k,d}^{(J+1)}}_{(d)}. \label{eqn:scale mapping module bound 1}
    \end{align}
    First note ${\displaystyle r_{k,1}^{(J)}\, \widehat{r}_{k,1}^{(J)} = r_{k,1}^{(J)}\prod_{j = 1}^{J} r_{k,1}^{(j-1)} = \prod_{j = 1}^{J+1} r_{k,1}^{(j-1)} = \widehat{r}_{k,1}^{(J+1)}}$.
    Second note
    \begin{align*}
        r_{k,1}^{(J)}\,\widehat{r}_{k,2}^{(J)} + r_{k,2}^{(J)} &= r_{k,1}^{(J)}\sum_{j = 1}^{J} r_{k,2}^{(j-1)}\prod_{\ell = j}^{J-1} r_{k,1}^{(\ell)} + r_{k,2}^{(J)}  \\
        &= \sum_{j = 1}^{J+1} r_{k,2}^{(j-1)}\prod_{\ell = j}^{J} r_{k,1}^{(\ell)} = \widehat{r}_{k,2}^{(J+1)}
    \end{align*}
    and similarly $r_{k,1}^{(J)}\,\widehat{r}_{k,d,3}^{(j, J)} = \widehat{r}_{k,d,3}^{(j, J+1)}$ and $r_{k,d,3}^{(J)} = \widehat{r}_{k,d,3}^{(J+1, J+1)}.$

 Combining these two notes with (\ref{eqn:scale mapping module bound 1}) gives the desired induction result.
\end{proof}

\subsection{Properties of the Tikhonov Solution}

Recall, for square matrix, $M$, the spectral norm, $\norm{M}_2$, is the largest absolute eigenvalue of the matrix. First, we provide a few lemmas necessary to derive a Lipschitz condition for the Tikhonov solution. 
\begin{lemma}\label{lemma:spectral norm bound on (AtA+P^-1)^-1}
    For any invertible symmetric matrix $P$ and scale variable $\bm{z}$ it holds that $\norm{(A_{\bm{z}}^TA_{\bm{z}}+P^{-1})^{-1}}_2 \leq \norm{P}_2.$
\end{lemma}
\begin{proof}
   Let $0 < \lambda_1\leq \cdots \leq \lambda_n$ be the eigenvalues of $A_{\bm{z}}^TA_{\bm{z}}+P^{-1}$, $0\leq \gamma_1\leq \cdots \leq \gamma_n$ be the eigenvalues of $A_{\bm{z}}^TA_{\bm{z}}$, and $0 < \kappa_1\leq \cdots \leq \kappa_n$ the eigenvalues of $P$. Note these eigenvalues are all non-negative as $A_{\bm{z}}^TA_{\bm{z}}$, $P$, and $A_{\bm{z}}^TA_{\bm{z}}+P^{-1}$  are real, symmetric matrices.  Then $0 < \lambda_n^{-1}\leq \cdots \leq \lambda_1^{-1}$ are the eigenvalues of $(A_{\bm{z}}^TA_{\bm{z}}+P^{-1})^{-1}$ and $0 < \kappa_n^{-1}\leq \cdots \leq \kappa_1^{-1}$ are the eigenvalues of $P^{-1}$. From Weyl's inequality
\begin{align*}
    \gamma_1 + \kappa_n^{-1} \leq \lambda_1
\end{align*}
and thus
\begin{align*}
  &\norm{(A_{\bm{z}}^TA_{\bm{z}}+P^{-1})^{-1}}_2 =  \lambda_1^{-1} \leq \frac{1}{\gamma_1 + \kappa_n^{-1}} \leq \kappa_n = \norm{P}_2. \qedhere
\end{align*}
\end{proof}

\begin{lemma} \label{lemma:spectal norm bound ||AtA(P2) - AtA(P1)||}
    Let $\bm{z}_1$ and $\bm{z}_2$ satisfy $\norm{\bm{z}_1}_{\infty}, \norm{\bm{z}_2}_{\infty}\leq z_{\infty}$. Then
    \begin{align*}
         \norm{A_{\bm{z}_2}^TA_{\bm{z}_2}-  A_{\bm{z}_1}^TA_{\bm{z}_1}}_2 & \leq  2z_{\infty} \norm{A}_2^2 \norm{\bm{z}_1 - \bm{z}_2}_{\infty}.
    \end{align*}
\end{lemma}
\begin{proof}
Observe
\begin{align*}
    &\norm{A_{\bm{z}_2}^TA_{\bm{z}_2}-  A_{\bm{z}_1}^TA_{\bm{z}_1}}_2 \\
    &=  \norm{A_{\bm{z}_2}^TA_{\bm{z}_2}-A_{\bm{z}_2}^TA_{\bm{z}_1} + A_{\bm{z}_2}^TA_{\bm{z}_1} -  A_{\bm{z}_1}^TA_{\bm{z}_1}}_2 \\
    &\leq \norm{A_{\bm{z}_2}^T\left(A_{\bm{z}_2} - A_{\bm{z}_1}\right)}_2 + \norm{\left(A_{\bm{z}_2}^T - A_{\bm{z}_1}^T\right)A_{\bm{z}_1}}_2 \\
    &\leq \left(\norm{\bm{z}_1}_{\infty}+\norm{\bm{z}_2}_{\infty}\right) \norm{A^TA}_2 \norm{\bm{z}_1 - \bm{z}_2}_{\infty} \\
    &\leq 2z_{\infty}\norm{A}_2^2 \norm{\bm{z}_1 - \bm{z}_2}_{\infty}. \qedhere
\end{align*}
\end{proof}

\begin{lemma}\label{lemma:spectral norm bound ||P1^-1 - P2^-1||}
    For any invertible matrices $P$ and $\widetilde{P}$ 
    \begin{align*}
        \norm{\widetilde{P}^{-1} - P^{-1}}_2 & \leq \norm{P^{-1}}_2 \norm{\widetilde{P}^{-1}}_2 \norm{P - \widetilde{P}}_2.
    \end{align*}
\end{lemma}
\begin{proof}
    Observe
    \begin{align*}
        \norm{\widetilde{P}^{-1} - P^{-1}}_2         &= \norm{\widetilde{P}^{-1}\left(P - \widetilde{P}\right) P^{-1}}_2  \\
        &\leq \norm{P^{-1}}_2 \norm{\widetilde{P}^{-1}}_2 \norm{P - \widetilde{P}}_2. \qedhere
    \end{align*}
\end{proof}

Next, we bound the spectral norm on the difference of two invertible portions of the Tikhonov solution. For an invertible matrix, $M$, recall that the condition number of $M$ is
\[
\textnormal{Cond}(M) = \norm{M}_2\norm{M^{-1}}_2.
\]

\begin{corollary}\label{corollary:spectral norm bound ||(AtA(P1)+P1^-1)^-1 - (AtA(P2)+P2^-1)^-1||}
    For any invertible symmetric matrices $P$ and $\widetilde{P}$ and any $\bm{z}_1$ and $\bm{z}_2$ satisfying $\norm{\bm{z}_1}_{\infty}, \norm{\bm{z}_2}_{\infty}\leq z_{\infty}$ it holds
    \begin{multline*}
    \norm{(A_{\bm{z}_1}^TA_{\bm{z}_1} + P^{-1})^{-1} - (A_{\bm{z}_2}^TA_{\bm{z}_2} + \widetilde{P}^{-1})^{-1}}_2 \\
 \leq 2z_{\infty}\norm{A}_2^2\norm{P}_2\norm{\widetilde{P}}_2 \norm{\bm{z}_1-\bm{z}_2}_{\infty} \\
     + \textnormal{Cond}(P)\textnormal{Cond}(\widetilde{P}) \norm{P - \widetilde{P}}_2.
    \end{multline*}
\end{corollary}
\begin{proof}
Using Lemma \ref{lemma:spectral norm bound ||P1^-1 - P2^-1||} and then Lemma \ref{lemma:spectral norm bound on (AtA+P^-1)^-1} note
\begin{multline*}
    \norm{(A_{\bm{z}_1}^TA_{\bm{z}_1} + P^{-1})^{-1} - (A_{\bm{z}_2}^TA_{\bm{z}_2} + \widetilde{P}^{-1})^{-1}}_2 \\
    \leq \norm{P}_2\norm{\widetilde{P}}_2\norm{A_{\bm{z}_2}^TA_{\bm{z}_2} + \widetilde{P}^{-1} - A_{\bm{z}_1}^TA_{\bm{z}_1} - P^{-1}}_2.
\end{multline*}
Next, using  Lemma \ref{lemma:spectal norm bound ||AtA(P2) - AtA(P1)||} and Lemma \ref{lemma:spectral norm bound ||P1^-1 - P2^-1||} observe
\begin{align*}
    &\norm{A_{\bm{z}_2}^TA_{\bm{z}_2} + \widetilde{P}^{-1} - A_{\bm{z}_1}^TA_{\bm{z}_1} - P^{-1}}_2 \\
    & \leq \norm{A_{\bm{z}_2}^TA_{\bm{z}_2} - A_{\bm{z}_1}^TA_{\bm{z}_1}}_2 + \norm{\widetilde{P}^{-1} - P^{-1}}_2 \\
    &\leq 2z_{\infty}\norm{A}_2^2 \norm{\bm{z}_1 - \bm{z}_2}_{\infty} + \norm{P^{-1}}_2\norm{\widetilde{P}^{-1}}_2 \norm{P - \widetilde{P}}_2. \qedhere
\end{align*}
\end{proof}

Now, we prove the Tikhonov solution is Lipschitz.

\begin{prop} \label{prop:lipschitz of u}
Let $\bm{z}_1$ and $\bm{z}_2$ satisfy $\norm{\bm{z}_1}_{\infty}, \norm{\bm{z}_2}_{\infty}\leq z_{\infty}$. Then the Tikhonov solution $\mathcal{T}_{\bm{y}}$, which is parameterized by a SPD matrix $P$, satisfies
\begin{align*}
   \resizebox{\columnwidth}{!}{${\displaystyle\norm{\mathcal{T}_{\bm{y}}(\bm{z}_1; P) - \mathcal{T}_{\bm{y}}(\bm{z}_2; \widetilde{P})}_2 \leq c_1(\bm{y}) \norm{\bm{z}_1-\bm{z}_2}_{\infty} + c_2(\bm{y}) \norm{P - \widetilde{P}}_2 }$}
\end{align*}
for
\begin{align*}
    &c_1(\bm{y}) = \norm{P}_2\norm{A}_2 \norm{\bm{y}}_2\left(1 + 2z_{\infty}^2\norm{\widetilde{P}}_2 \norm{A}_2^2\right) \\
    &c_2(\bm{y}) = z_{\infty} \norm{A}_2 \norm{\bm{y}}_2 \,\, \textnormal{Cond}(P)\textnormal{Cond}(\widetilde{P}).
\end{align*}
\end{prop}
\begin{proof}
    Observe
\begin{align*}
    &\norm{\mathcal{T}_{\bm{y}}(\bm{z}_1; P) - \mathcal{T}_{\bm{y}}(\bm{z}_2; \widetilde{P})}_2 \\
    &= \norm{(A_{\bm{z}_1}^TA_{\bm{z}_1} + P^{-1})^{-1}A_{\bm{z}_1}^T\bm{y} - (A_{\bm{z}_2}^TA_{\bm{z}_2} + \widetilde{P}^{-1})^{-1}A_{\bm{z}_2}^T\bm{y} }_2 \\
    &= \left\lVert(A_{\bm{z}_1}^TA_{\bm{z}_1} + P^{-1})^{-1}A_{\bm{z}_1}^T\bm{y} - (A_{\bm{z}_1}^TA_{\bm{z}_1} + P^{-1})^{-1}A_{\bm{z}_2}^T\bm{y} \right. \\
    &\hspace{.3cm} \left. + (A_{\bm{z}_1}^TA_{\bm{z}_1} + P^{-1})^{-1}A_{\bm{z}_2}^T\bm{y}- (A_{\bm{z}_2}^TA_{\bm{z}_2} +\widetilde{P}^{-1})^{-1}A_{\bm{z}_2}^T\bm{y}\right\rVert_2 \\
    &= \left\lVert(A_{\bm{z}_1}^TA_{\bm{z}_1} + P^{-1})^{-1}\left(A_{\bm{z}_1}^T - A_{\bm{z}_2}^T\right)\bm{y} \right.\\
    &\hspace{.3cm} \left. + \left[(A_{\bm{z}_1}^TA_{\bm{z}_1} + P^{-1})^{-1} - (A_{\bm{z}_2}^TA_{\bm{z}_2} + \widetilde{P}^{-1})^{-1}\right]A_{\bm{z}_2}^T\bm{y} \right\rVert_2 \\
    &\leq \norm{(A_{\bm{z}_1}^TA_{\bm{z}_1} + P^{-1})^{-1}}_2 \norm{\bm{y}}_2 \norm{A_{\bm{z}_1}^T - A_{\bm{z}_2}^T}_2 +\\
    & \resizebox{\columnwidth}{!}{${\displaystyle z_{\infty} \norm{A}_2\norm{\bm{y}}_2 \norm{(A_{\bm{z}_1}^TA_{\bm{z}_1} + P^{-1})^{-1} - (A_{\bm{z}_2}^TA_{\bm{z}_2} + \widetilde{P}^{-1})^{-1}}_2}$} \\
    &\leq \norm{P}_2\norm{A}_2 \norm{\bm{y}}_2\,\, \norm{\bm{z}_1 - \bm{z}_2}_{\infty} +\\
    & \resizebox{\columnwidth}{!}{${\displaystyle z_{\infty} \norm{A}_2\norm{\bm{y}}_2 \norm{(A_{\bm{z}_1}^TA_{\bm{z}_1} + P^{-1})^{-1} - (A_{\bm{z}_2}^TA_{\bm{z}_2} + \widetilde{P}^{-1})^{-1}}_2}$}
\end{align*}
where in the last inequality we used Lemma \ref{lemma:spectral norm bound on (AtA+P^-1)^-1}. Combining the above inequality with Corollary \ref{corollary:spectral norm bound ||(AtA(P1)+P1^-1)^-1 - (AtA(P2)+P2^-1)^-1||} produces the desired result.
\end{proof}

Lastly, we derive a bound on the Tikhonov solution

\begin{prop} \label{prop:u bound}
    For any SPD matrix $P$, any $\bm{z}$, $\bm{y}$, and $2\leq p\leq \infty$ it holds that $\norm{\mathcal{T}_{\bm{y}}(\bm{z};P)}_p \leq \norm{\bm{z}}_{\infty}\norm{P}_2\norm{A}_p\norm{\bm{y}}_p.$
\end{prop}
\begin{proof}
Using Lemma  \ref{lemma:spectral norm bound on (AtA+P^-1)^-1} observe
    \begin{align*}
    \norm{\mathcal{T}_{\bm{y}}(\bm{z};P)}_p &= \norm{(A_{\bm{z}}^TA_{\bm{z}} + P^{-1})^{-1}A_{\bm{z}}^T\bm{y}}_p \\
    &\leq \norm{(A_{\bm{z}}^TA_{\bm{z}} + P^{-1})^{-1}}_p\norm{\textnormal{Diag}(\bm{z})}_p \norm{A}_p  \norm{\bm{y}}_p \\
    &\leq \norm{(A_{\bm{z}}^TA_{\bm{z}} + P^{-1})^{-1}}_2\norm{\textnormal{Diag}(\bm{z})}_2 \norm{A}_p  \norm{\bm{y}}_p \\
    &\leq \norm{\bm{z}}_{\infty}\norm{P}_2\norm{A}_p\norm{\bm{y}}_p. \qedhere
\end{align*}
\end{proof}

\subsection{Lipschitz Property of Network Outputs}
In this section, we define $\bm{\zeta}_k$ and $\widetilde{\bm{\zeta}}_k$ as the G-CG-Net scale variable estimates on iteration $k$ when G-CG-Net is parameterized by $\bm{\Theta}$ or $\widetilde{\bm{\Theta}}$, from (\ref{eqn:G-CG-Net params}), respectively. That is, $\bm{\zeta}_k$ is recursively defined by $\bm{\zeta}_k = \mathcal{Z}_k^{(J)}(\bm{\zeta}_{k-1}, \mathcal{T}_{\bm{y}}(\bm{\zeta}_{k-1};P_u);\Theta_k^{(J)})$

First, we show a Lipschitz property of the final scale variable estimate in the following proposition.

\begin{prop} \label{prop:scale variable lipschitz z_K}
    If Assumption \ref{assumption:scale update method} holds then
    \begin{multline*}
        \norm{\bm{\zeta}_K - \widetilde{\bm{\zeta}}_K}_2 \leq \widehat{c}_1^{(K,J)}(\bm{y}) \norm{P_u - \widetilde{P}_u}_2 \\
        + \sum_{k = 1}^K\sum_{j = 1}^J\sum_{d = 1}^D \widehat{c}_{k, j, d, 2}^{(K,J)}(\bm{y}) \norm{\bm{\theta}_{k, d}^{(j)} - \widetilde{\bm{\theta}}_{k,d}^{(j)}}_{(d)}
    \end{multline*}
    for
    \begin{align*}
        \widehat{c}_1^{(K,J)}(\bm{y}) &= c_2(\bm{y})\sum_{k = 1}^K \widehat{r}_{k,2}^{(J)} \prod_{\ell = k+1}^K (\widehat{r}_{\ell,1}^{(J)}+\widehat{r}_{\ell,2}^{(J)} c_1(\bm{y})) \\
        \widehat{c}_{k, j, d, 2}^{(K,J)}(\bm{y}) &= \widehat{r}_{k,d,3}^{(j, J)}\prod_{\ell = k+1}^K (\widehat{r}_{\ell,1}^{(J)}+\widehat{r}_{\ell,2}^{(J)} c_1(\bm{y}))
    \end{align*}
    where $\widehat{r}_{k,1}^{(J)}, \widehat{r}_{k,2}^{(J)}$, and $\widehat{r}_{k,d,3}^{(j, J)}$ are given in Proposition \ref{prop:scale variable lipschitz 1} and $c_1(\bm{y})$ and $c_2(\bm{y})$ are given in Proposition \ref{prop:lipschitz of u}.
\end{prop}
\begin{proof}
    Combining Proposition \ref{prop:scale variable lipschitz 1} and \ref{prop:lipschitz of u} it holds for any $k$
    \begin{align}
        &\norm{\bm{\zeta}_k - \widetilde{\bm{\zeta}}_k}_2 \nonumber \\
        &= \left\lVert\mathcal{Z}_k^{(J)}(\bm{\zeta}_{k-1},\mathcal{T}_{\bm{y}}(\bm{\zeta}_{k-1}; P_u); \Theta_k^{(J)}) \right.\nonumber \\
        &\hspace{.45cm} \left.- \mathcal{Z}_k^{(J)}(\widetilde{\bm{\zeta}}_{k-1},\mathcal{T}_{\bm{y}}(\widetilde{\bm{\zeta}}_{k-1}; \widetilde{P}_u); \widetilde{\Theta}_k^{(J)})\right\rVert_2 \nonumber \\
        &\resizebox{\columnwidth}{!}{${\displaystyle \leq (\widehat{r}_{k,1}^{(J)}+\widehat{r}_{k,2}^{(J)} c_1(\bm{y})) \norm{\bm{\zeta}_{k-1}-\widetilde{\bm{\zeta}}_{k-1}}_2 + \widehat{r}_{k,2}^{(J)}c_2(\bm{y}) \norm{P_u -\widetilde{P}_u}_2 }$} \nonumber \\
        &\hspace{.45cm} + \sum_{j = 1}^J\sum_{d = 1}^D \widehat{r}_{k,d,3}^{(j, J)} \norm{\bm{\theta}_{k, d}^{(j)} - \widetilde{\bm{\theta}}_{k,d}^{(j)}}_{(d)}. \label{eqn:scale variable lipschitz bound k}
    \end{align}
    Now, we use induction on $K$. The base case $K = 1$ holds by (\ref{eqn:scale variable lipschitz bound k}) as $\bm{\zeta}_0 = \widetilde{\bm{\zeta}}_0 \equiv \mathcal{Z}_0$ and $\prod_{\ell = K+1}^K (\widehat{r}_{\ell,1}^{(J)}+\widehat{r}_{\ell,2}^{(J)} c_1(\bm{y})) = 1$.    
    Assume the induction hypothesis holds for fixed $K-1$ where $K > 2$. By (\ref{eqn:scale variable lipschitz bound k}) and the induction hypothesis observe
    \begin{align}
        &\norm{\bm{\zeta}_{K} - \widetilde{\bm{\zeta}}_{K}}_2 \nonumber \leq (\widehat{r}_{K,1}^{(J)}+\widehat{r}_{K,2}^{(J)} c_1(\bm{y})) \norm{\bm{\zeta}_{K-1}-\widetilde{\bm{\zeta}}_{K-1}}_2 \\
        &\hspace{.45cm}+ \widehat{r}_{K,2}^{(J)}c_2(\bm{y}) \norm{P_u -\widetilde{P}_u}_2 \nonumber  \\
        &\hspace{.45cm} + \sum_{j = 1}^J\sum_{d = 1}^D \widehat{r}_{K,d,3}^{(j, J)} \norm{\bm{\theta}_{K, d}^{(j)} - \widetilde{\bm{\theta}}_{K,d}^{(j)}}_{(d)} \nonumber \\
        &\resizebox{\columnwidth}{!}{${\displaystyle \leq \left[(\widehat{r}_{K,1}^{(J)}+\widehat{r}_{K,2}^{(J)} c_1(\bm{y}))\widehat{c}_1^{(K-1,J)}(\bm{y})  + \widehat{r}_{K,2}^{(J)}c_2(\bm{y})\right] \norm{P_u - \widetilde{P}_u}_2}$} \nonumber \\
        &\resizebox{\columnwidth}{!}{${\displaystyle+ (\widehat{r}_{K,1}^{(J)}+\widehat{r}_{K,2}^{(J)} c_1(\bm{y}))\sum_{k = 1}^{K-1}\sum_{j = 1}^J\sum_{d = 1}^D \widehat{c}_{k, j, d, 2}^{(K-1,J)}(\bm{y}) \norm{\bm{\theta}_{k, d}^{(j)} - \widetilde{\bm{\theta}}_{k,d}^{(j)}}_{(d)}}$} \nonumber \\
        &\hspace{.45cm} + \sum_{j = 1}^J\sum_{d = 1}^D \widehat{r}_{K,d,3}^{(j, J)} \norm{\bm{\theta}_{K, d}^{(j)} - \widetilde{\bm{\theta}}_{K,d}^{(j)}}_{(d)}. \label{eqn:scale variable lipschitz induction}
    \end{align}
    First note,
    \begin{align*}
        &(\widehat{r}_{K,1}^{(J)}+\widehat{r}_{K,2}^{(J)} c_1(\bm{y}))\widehat{c}_1^{(K-1,J)}(\bm{y})  + \widehat{r}_{K,2}^{(J)}c_2(\bm{y}) \\
        &\resizebox{\columnwidth}{!}{${\displaystyle =(\widehat{r}_{K,1}^{(J)}+\widehat{r}_{K,2}^{(J)} c_1(\bm{y})) c_2(\bm{y})\sum_{k = 1}^{K-1} \widehat{r}_{k,2}^{(J)} \prod_{\ell = k+1}^{K-1} (\widehat{r}_{\ell,1}^{(J)}+\widehat{r}_{\ell,2}^{(J)} c_1(\bm{y}))}$} \\
        &\hspace{.45cm} + \widehat{r}_{K,2}^{(J)}c_2(\bm{y}) \\
        &= c_2(\bm{y})\sum_{k = 1}^{K-1} \widehat{r}_{k,2}^{(J)} \prod_{\ell = k+1}^{K} (\widehat{r}_{\ell,1}^{(J)}+\widehat{r}_{\ell,2}^{(J)} c_1(\bm{y})) \\
        &\hspace{.45cm} + \widehat{r}_{K,2}^{(J)}c_2(\bm{y}) \prod_{\ell = K+1}^{K} (\widehat{r}_{\ell,1}^{(J)}+\widehat{r}_{\ell,2}^{(J)} c_1(\bm{y})) \\
        &= c_2(\bm{y})\sum_{k = 1}^{K} \widehat{r}_{k,2}^{(J)} \prod_{\ell = k+1}^{K} (\widehat{r}_{\ell,1}^{(J)}+\widehat{r}_{\ell,2}^{(J)} c_1(\bm{y})) \\
        &= \widehat{c}_1^{(K,J)}(\bm{y}).
    \end{align*}
    Similarly, note
    \begin{align*}
        (\widehat{r}_{K,1}^{(J)}+\widehat{r}_{K,2}^{(J)} c_1(\bm{y}))\widehat{c}_{k, j, d, 2}^{(K-1,J)}(\bm{y}) = \widehat{c}_{k, j, d, 2}^{(K,J)}(\bm{y})
    \end{align*}
    and
    \begin{align*}
        \widehat{r}_{K,d,3}^{(j, J)} &= \widehat{r}_{K,d,3}^{(j, J)}\prod_{\ell = K+1}^{K} (\widehat{r}_{\ell,1}^{(J)}+\widehat{r}_{\ell,2}^{(J)} c_1(\bm{y})) \\
        &= \widehat{c}_{K, j, d, 2}^{(K,J)}(\bm{y}).
    \end{align*}
    Combining these two notes with (\ref{eqn:scale variable lipschitz induction}) produces the desired result.
\end{proof}

Finally, we show that G-CG-Net estimates, $\widehat{\bm{c}}(\bm{y};\bm{\Theta})$, are Lipschitz w.r.t. the G-CG-Net parameters.
\begin{theorem} \label{thm:G-CG-Net Lipschitz}
    Let Assumption \ref{assumption:scale update method} hold. Then for any parameterizations  $\bm{\Theta}$ and $\widetilde{\bm{\Theta}}$ of G-CG-Net, the G-CG-Net estimates satisfy the following Lipschitz property:
    \begin{align*}
        &\norm{\widehat{\bm{c}}(\bm{y};\bm{\Theta}) - \widehat{\bm{c}}(\bm{y};\widetilde{\bm{\Theta}})}_2 \\
        &\leq \kappa(\bm{y}) \norm{P_u - \widetilde{P}_u}_2 + \sum_{k = 1}^K\sum_{j = 1}^J\sum_{d = 1}^D \kappa_{k,d}^{(j)}(\bm{y}) \norm{\bm{\theta}_{k, d}^{(j)} - \widetilde{\bm{\theta}}_{k,d}^{(j)}}_{(d)}
    \end{align*}
    for 
    \begin{align*}
        &\resizebox{\columnwidth}{!}{${\displaystyle \kappa(\bm{y}) = z_{\infty}(c_1(\bm{y})+\norm{P_u}_2\norm{A}_{\infty}\norm{\bm{y}}_{\infty})\widehat{c}_1^{(K,J)}(\bm{y}) + z_{\infty} c_2(\bm{y}) }$} \\
        &\kappa_{k,d}^{(j)}(\bm{y}) =  z_{\infty}(c_1(\bm{y})+\norm{P_u}_2\norm{A}_{\infty}\norm{\bm{y}}_{\infty})\widehat{c}_{k,j,d,2}^{(K,J)}(\bm{y})
    \end{align*}
    where $\widehat{c}_1^{(K,J)}(\bm{y}), \hspace{.25cm} \widehat{c}_{k,j,d,2}^{(K,J)}(\bm{y})$ are given in Proposition \ref{prop:scale variable lipschitz z_K} and $c_1(\bm{y}), c_2(\bm{y})$ are given in Proposition \ref{prop:lipschitz of u}.
\end{theorem}
\begin{proof}
    As $\rho_{c_{\max}}$ is $1$-Lipschitz observe
    \begin{align*}
        &\norm{\widehat{\bm{c}}(\bm{y};\bm{\Theta}) - \widehat{\bm{c}}(\bm{y};\widetilde{\bm{\Theta}})}_2 \\
        &= \norm{\rho_{c_{\max}}(\bm{\zeta}_K\circ \mathcal{T}_{\bm{y}}(\bm{\zeta}_K; P_u)) - \rho_{c_{\max}}(\widetilde{\bm{\zeta}}_K\circ \mathcal{T}_{\bm{y}}(\widetilde{\bm{\zeta}}_K; \widetilde{P}_u))}_2 \\
        &\leq \left\lVert\bm{\zeta}_K\circ \mathcal{T}_{\bm{y}}(\bm{\zeta}_K; P_u) - \widetilde{\bm{\zeta}}_K\circ \mathcal{T}_{\bm{y}}(\bm{\zeta}_K; P_u) \right.\\
        &\hspace{.65cm}\left.+\widetilde{\bm{\zeta}}_K\circ \mathcal{T}_{\bm{y}}(\bm{\zeta}_K; P_U)- \widetilde{\bm{\zeta}}_K\circ \mathcal{T}_{\bm{y}}(\widetilde{\bm{\zeta}}_K; \widetilde{P}_u)\right\rVert_2 \\
        &\leq \norm{\mathcal{T}_{\bm{y}}(\bm{\zeta}_K; P_u)}_{\infty} \norm{\bm{\zeta}_K - \widetilde{\bm{\zeta}}_K}_2 \\
        &\hspace{.65cm}+ z_{\infty} \norm{\mathcal{T}_{\bm{y}}(\bm{\zeta}_K; P_u) - \mathcal{T}_{\bm{y}}(\widetilde{\bm{\zeta}}_K; \widetilde{P}_u)}_2.
    \end{align*}
    Using Proposition \ref{prop:lipschitz of u}, \ref{prop:u bound}, and \ref{prop:scale variable lipschitz z_K} produces the final result.
\end{proof}

\section{Conclusion and Future Work}

For a class of compound Gaussian prior based DNNs that solve linear inverse problems, a generalization error bound was derived. Subsequently, this generalization error bound was applied to two realizations, namely, CG-Net and DR-CG-Net, showing bounds for these cases. The developed generalization error bound was produced by bounding the Rademacher complexity of the network hypothesis class by Dudley's inequality, which is further bounded using a Lipschitz property to estimate covering numbers of the network hypothesis class. A key contribution was in showing the parameters of compound Gaussian DNNs satisfy a Lipschitz condition under reasonable assumptions thereby allowing us to produce generalization bounds for CG-based DNNs.

While the derived generalization error shows with sufficient training data, roughly scaling quadratically in signal dimension and cubically in network size, small generalization guarantees can be met, it still remains to be shown that such a property is true for small training datasets. This is desirable as CG-Net and DR-CG-Net significantly outperform comparative methods in image estimation problems when trained only on a small dataset. Likely, aspects of the iterative algorithm, in particular the Tikhonov solution, would need to be further leveraged to provide insight into a small training generalization error bound. Furthermore, PAC-Bayes generalization bounds~\cite{alquier2021PACBayes} for CG-based DNNs, extending the derived bounds in this paper, is another open question that can provide greater insights into the generalization for low training scenarios.

\section{Appendix: Preliminaries} \label{apndx:proofs}

\subsection{Rademacher Complexity} \label{apndx:rademacher complexity}

Recall a Rademacher variable is a discrete random variable $\gamma$ taking values $\pm 1$ with equal probability.
\begin{deff}
    Let $\mathcal{G}$ be a set of functions $g:\mathcal{X}\to\mathbb{R}$  and $\mathcal{S} = \{\bm{x}_i\}_{i = 1}^{N_s} \subseteq \mathcal{X}$. The \textit{\textbf{empirical Rademacher complexity}} is
    \begin{align*}
        \mathcal{R}_{\mathcal{S}}(\mathcal{G}) = \mathbb{E}_{\bm{\gamma}}\left[\sup_{g\in\mathcal{G}} \frac{1}{N_s}\sum_{i = 1}^{N_s}\gamma_i g(\bm{x}_i)\right]
    \end{align*}
    for $\bm{\gamma} = [\gamma_1, \ldots, \gamma_{N_s}]$ a vector of i.i.d. Rademacher variables.
\end{deff}

For a DNN with hypothesis space $\mathcal{H}$ and using loss function $L$, we are interested in the Rademacher complexity of the real-valued set of functions $\mathcal{G} = L\circ \mathcal{H} = \{L\circ h : h\in\mathcal{H}\}.$

\begin{theorem}[\hspace*{-3px}\cite{shalev2014understanding}, Theorem 26.5] \label{thm:GEB by Rademacher complexity}
Let $\mathcal{H}$ be a set of functions, $\mathcal{S} = \{(\overline{\bm{y}}_i,\overline{\bm{c}}_i)\}_{i = 1}^{N_s}$ a training set drawn i.i.d. from $\mathcal{D}$, and $L$ a real-valued loss function satisfying $|L(h(\bm{y}),\bm{c})| \leq c$ for all $h\in\mathcal{H}$ and $(\bm{y},\bm{c})\sim \mathcal{D}$. Then, for $\varepsilon\in (0,1)$ with probability at least $1-\varepsilon$ we have for all $h\in\mathcal{H}$
    \begin{align*}
        \mathcal{L}(h)\leq \mathcal{L}_{\mathcal{S}}(h) + 2\mathcal{R}_{\mathcal{S}}(L\circ \mathcal{H}) + 4c\sqrt{2\ln(4/\varepsilon)/N_s}.
    \end{align*}
\end{theorem}
Next, we provide a contraction lemma allowing us to ignore the loss function and only consider the hypothesis class.
\begin{lemma}
[\hspace*{-3px}\cite{maurer2016vectorcontraction}, Corollary 4] \label{lemma:contraction lemma} Let $\mathcal{H}$ be a set of functions $h:\mathcal{X}\to\mathbb{R}^d$ and $\mathcal{S} = \{\bm{x}_i\}_{i = 1}^{N_s}\subseteq \mathcal{X}$. Then for any $\tau$-Lipschitz functions $g_i:\mathbb{R}^d\to\mathbb{R}$, where $i \in\mathbb{N}[N_s]$,
    \begin{align*}
       \resizebox{\columnwidth}{!}{${\displaystyle\mathbb{E}_{\bm{\gamma}}\left[\sup_{h\in\mathcal{H}}\sum_{i = 1}^{N_s} \gamma_i g_i\circ h(\bm{x}_i)\right] \leq \sqrt{2}\tau \mathbb{E}_{\Gamma}\left[\sup_{h\in\mathcal{H}} \sum_{i = 1}^{N_s}\sum_{k = 1}^d \gamma_{ik} h_k(\bm{x}_i)\right]}$}
    \end{align*}
    where $\bm{\gamma} = [\gamma_i]_{i \in\mathbb{N}[N_s]}$ and $\Gamma = [\gamma_{ik}]_{i \in\mathbb{N}[N_s]}^{k \in\mathbb{N}[d]}$ are collections of i.i.d. Rademacher variables.
\end{lemma}

\subsection{Dudley's Inequality} \label{apndx:Dudley's inequality}

To define Dudley's inequality, which is employed to obtain a bound on $\mathcal{R}_{\mathcal{S}}(L\circ \mathcal{H})$ in Theorem \ref{thm:GEB by Rademacher complexity}, we require some terminology from~\cite{devroye2013probabilistic, Compressive_Sensing}. First, for a metric space $(M, d)$, let $C_{M}(\epsilon)\subseteq M$ denote an \textbf{$\epsilon$-covering} of $M$. That is, for every $x\in M$ there exists a $c\in C_{M}(\epsilon)$ such that $d(x, c)\leq \epsilon.$ Second, let $\mathcal{N}(M, d, \epsilon)\in\mathbb{R}$ denote the \textbf{$\epsilon$-covering number} of $M$. That is, $\mathcal{N}(M, d, \epsilon)$ is the minimum cardinality of all $C_{M}(\epsilon)$ or, equivalently, the minimum number of $\epsilon$ radius balls as measured by $d$, to contain $M$. Third, a real-valued stochastic process $(X_t)_{t\in T}$ is called a \textbf{subgaussian process} if $\mathbb{E}(X_t) = 0$ and for all $s\in T$, $t\in T$ and $\theta > 0$
\begin{align*}
    \mathbb{E}\left[\exp(\theta(X_s - X_t))\right] \leq \exp\left(\theta^2\widetilde{d}(X_s,X_t)^2/2\right)
\end{align*}
where $\widetilde{d}(X_s,X_t) =(\mathbb{E}\left|X_s - X_t\right|^2)^{1/2}$ is a pseudo-metric. Fourth, the radius of $T$ is $\Delta(T) = \sup_{t\in T} \sqrt{\mathbb{E}|X_t|^2}$.

\begin{lemma}[Dudley's Inequality, \cite{devroye2013probabilistic}, Theorem 8.23~\cite{Compressive_Sensing}] \label{lemma:dudley's inequality}
    For a subgaussian stochastic process $(X_t)_{t\in T}$ with pseudo-metric $\widetilde{d}$
    \begin{align*}
\mathbb{E}\left(\sup_{t\in T} X_t\right) \leq 4\sqrt{2}\int_0^{\Delta(T)/2} \sqrt{\ln\left(\mathcal{N}(T, \widetilde{d}, \epsilon)\right)} d\epsilon.
    \end{align*}
\end{lemma}

\subsection{Rademacher Process} \label{apndx:Rademacher process}

Let $V$ be a vector space and $\mathcal{X} = V^T = \{f:T \to V\}$. A \textbf{Rademacher process}, $(Y_t)_{t\in T}$, is a stochastic process of the form $ Y_t = \sum_{k = 1}^n \gamma_{k} x_k(t)$ where $x_k\in\mathcal{X}$ and $\bm{\gamma} = [\gamma_1, \ldots, \gamma_n]$ are i.i.d. Rademacher variables. As $\mathbb{E}(\gamma_k) = 0$, $\gamma_k^2 = 1$, and $\gamma_k, \gamma_j$ are independent for $k\neq j$, then for real-valued Rademacher processes, i.e. $V = \mathbb{R}$,
\begin{align}
    \mathbb{E}_{\bm{\gamma}}|Y_t|^2 &= \sum_{k = 1}^n\mathbb{E}_{\bm{\gamma}}\left( \gamma_k^2 x_k(t)^2\right) + \sum_{k = 1}^n\sum_{\substack{j = 1 \\ j\neq k}}^n\mathbb{E}_{\bm{\gamma}}\left(\gamma_j\gamma_k x_j(t)x_k(t)\right) \nonumber \\
    &= \sum_{k = 1}^n x_k(t)^2. \label{eqn:expectation of Rademacher process squared}
\end{align}
Similarly,
\begin{align}
    \widetilde{d}(Y_s,Y_t)^2 &= \mathbb{E}_{\bm{\gamma}}|Y_s-Y_t|^2 = \sum_{k = 1}^n (x_k(s)-x_k(t))^2. \label{eqn:pseudometric}
\end{align}
Finally, we remark that real-valued Rademacher processes are subgaussian~\cite{Compressive_Sensing} and thus satisfy Lemma \ref{lemma:dudley's inequality}.

\subsection{Covering Number Bounds}

First, a covering number bound on a subset of the unit ball.
\begin{lemma}[\hspace*{-3px}\cite{devroye2013probabilistic}, Proposition C.3~\cite{Compressive_Sensing}] \label{lemma:covering number bound on unit subset}
    For any norm $\norm{\cdot}$ on $\mathbb{R}^n$ and subset $U\subseteq \{\bm{x}\in\mathbb{R}^n: \norm{\bm{x}}\leq 1\}$ it holds
    \begin{align*}
        \mathcal{N}(U, \norm{\cdot}, \epsilon) \leq \left(1+2/\epsilon\right)^n.
    \end{align*}
\end{lemma}
Second, a covering number bound on a space of parametric functions satisfying a Lipschitz criterion.
\begin{lemma} \label{lemma:covering number bound on parametric function}
    Let $(\Theta, \norm{\cdot}_{\vartheta})$ be a non-empty, bounded, and normed vector space. Define $\mathcal{F} = \{f_{\theta}:\mathcal{X}\to\mathcal{Y}\,|\, \theta\in \Theta\}$ and let $\norm{\cdot}_{\mathcal{F}}$ be any norm on $\mathcal{F}$. Assume that for any $\theta,\widetilde{\theta}\in\Theta$
    \[
    \norm{f_{\theta}(\bm{x}) - f_{\widetilde{\theta}}(\bm{x})}_{\mathcal{F}} \leq \Gamma(\bm{x}) \norm{\theta-\widetilde{\theta}}_{\vartheta}
    \]
     and let $\Gamma = \sup_{\bm{x}\in\mathcal{X}} \Gamma(\bm{x})$. Then 
    \[
    \mathcal{N}(\mathcal{F}, \norm{\cdot}_{\mathcal{F}}, \epsilon) \leq \mathcal{N}(\Theta, \norm{\cdot}_{\vartheta}, \epsilon/\Gamma).
    \]
\end{lemma}
\begin{proof}

Let $C_{\Theta}$ be any $\epsilon/\Gamma$-covering of $\Theta$ and $C_{\mathcal{F}} = \{f_{\theta}: \theta\in C_{\Theta}\}.$ Note $|C_{\mathcal{F}}|\leq |C_{\Theta}|$ as, at most, every $\theta\in C_{\Theta}$ uniquely defines a function $f_\theta\in C_{\mathcal{F}}$. Further, for any $f_{\theta}\in\mathcal{F}$ there exists a $c\in C_{\Theta}$ such that $\norm{\theta-c}\leq \epsilon/\Gamma$. Thus $f_c\in C_{\mathcal{F}}$ satisfies
    \begin{align*}
        \norm{f_{\theta}(\bm{x}) - f_{c}(\bm{x})}_{\mathcal{F}} \leq \Gamma(\bm{x}) \norm{\theta - c}_{\vartheta} \leq \Gamma \norm{\theta - c}_{\vartheta} \leq \epsilon.
    \end{align*}
    Hence, every $\epsilon/\Gamma$-covering of $\Theta$ generates an $\epsilon$-covering of $\mathcal{F}$. Take $C_{\Theta}$ to be the minimal cardinality $\epsilon/\Gamma$-covering of $\Theta$, then  $\mathcal{N}(\mathcal{F}, \norm{\cdot}_{\mathcal{F}}, \epsilon) \leq |C_{\mathcal{F}}| = |C_{\Theta}| =  \mathcal{N}(\Theta, \norm{\cdot}_{\vartheta}, \epsilon/\Gamma)$.
\end{proof}

Finally, we generalize Lemma \ref{lemma:covering number bound on parametric function} to a class of functions parameterized by a sequence of parameters.
\begin{corollary} \label{corollary:covering number bound for multiple Lipschitz bounds}
    Let $(\Theta_i, \norm{\cdot}_{\vartheta_i})$, for $i \in \mathbb{N}[n]$, be a sequence of non-empty, bounded, and normed vector spaces. Define $\mathcal{F} = \{f_{\theta_1,\ldots, \theta_n}:\mathcal{X}\to\mathcal{Y}\,|\, \theta_i\in \Theta_i\}$ and let $\norm{\cdot}_{\mathcal{F}}$ be a norm on $\mathcal{F}$. Assume that
    \begin{align*}
        \norm{f_{\theta_1,\ldots, \theta_n}(\bm{x}) - f_{\widetilde{\theta}_1,\ldots, \widetilde{\theta}_n}(\bm{x})}_{\mathcal{F}} \leq \sum_{i = 1}^n \Gamma_i(\bm{x}) \norm{\theta_i - \widetilde{\theta}_i}_{\vartheta_i}
    \end{align*}
    and let $\Gamma_i = \sup_{\bm{x}\in\mathcal{X}} \Gamma_i(\bm{x})$. Then
    \begin{align*}
        \mathcal{N}(\mathcal{F}, \norm{\cdot}_{\mathcal{F}}, \epsilon)\leq \prod_{i = 1}^n \mathcal{N}(\Theta_i, \norm{\cdot}_{\vartheta_i}, \epsilon/(n\Gamma_i)).
    \end{align*}
\end{corollary}
\begin{proof}

    Let $(S_k, d_k)$, for $k \in\mathbb{N}[n]$, be metric spaces and define the product metric space $(\mathcal{S}, d)$ where $\mathcal{S} \coloneqq S_1\times \cdots\times S_n$ and $d(\bm{s}, \bm{c}) \coloneqq \sum_{k = 1}^n a_kd_k(s_k, c_k)$ for $\bm{s},\bm{c}\in\mathcal{S}$ and fixed $a_k\in (0,\infty)$. Let $C_k$ be any $\epsilon/(a_k n)$-covering of $S_k$ and define $\mathcal{C}\coloneqq C_1\times\cdots\times C_n$. Then for any $\bm{s}\in\mathcal{S}$ there exists a $\bm{c} \in \mathcal{C}$ such that
    $d(\bm{s},\bm{c}) = \sum_{k = 1}^n a_kd_k(s_k, c_k) \leq \sum_{k = 1}^n a_k\epsilon/(n a_k) = \epsilon,$ which implies that $\mathcal{C}$ is an $\epsilon$-covering of $\mathcal{S}$. Take every $C_k$ to be the minimal cardinality $\epsilon/(a_k n)$-covering of $S_k$ then
    \begin{align}
        \mathcal{N}(\mathcal{S}, d, \epsilon) \leq |\mathcal{C}| =  \prod_{k = 1}^n |C_k| = \prod_{k = 1}^n \mathcal{N}\left(S_k, d_k, \frac{\epsilon}{a_k n}\right). \label{eqn:covering number bound on product space}
    \end{align}

    Next, let $(\Theta, \norm{\cdot}_{\Theta})$ be the normed vector space where $\Theta = \Theta_1\times \cdots \times \Theta_n$ and for $\bm{\theta} = (\theta_1, \ldots, \theta_n)\in\Theta$ we define $\norm{\bm{\theta}}_{\Theta} = \sum_{i =1}^n \Gamma_i \norm{\theta_i}_{\vartheta_i}$. Then for any $\bm{\theta}, \widetilde{\bm{\theta}}\in\Theta$ where $\bm{\theta} = (\theta_1, \ldots, \theta_n)$ and $\widetilde{\bm{\theta}} =  (\widetilde{\theta}_1, \ldots, \widetilde{\theta}_n)$ we have
    \begin{align*}
        &\norm{f_{\theta_1,\ldots, \theta_n}(\bm{x}) - f_{\widetilde{\theta}_1,\ldots, {\theta}_n}(\bm{x})}_{\mathcal{F}} \\
        &\leq \sum_{i = 1}^n \Gamma_i(\bm{x}) \norm{\theta_i - \widetilde{\theta}_i}_{\vartheta_i} \leq \sum_{i = 1}^n \Gamma_i \norm{\theta_i - \widetilde{\theta}_i}_{\vartheta_i} = \norm{\bm{\theta} - \widetilde{\bm{\theta}}}_{\Theta}.
    \end{align*}
    Combining this Lipschitz property with Lemma \ref{lemma:covering number bound on parametric function} and then applying (\ref{eqn:covering number bound on product space}) produces the covering number bound.
\end{proof}

\section{Appendix: Proof of Theorem \ref{thm:GEB G-CG-Net}} \label{apndx:G-CG-Net GEB proof}

Proving Theorem \ref{thm:GEB G-CG-Net} consists of the following steps: \textbf{Step (1)}, we establish a Rademacher process generated by $\mathcal{H}_{\textnormal{CG}}^{(1)}$. \textbf{Step (2)}, using Lemma \ref{lemma:contraction lemma}, we express the Rademacher complexity of $L\circ \mathcal{H}_{\textnormal{CG}}^{(1)}$ as the expected supremum of Step (1). \textbf{Step (3)}, invoking Dudley's inequality we bound Step (2) by an integral of some covering numbers. \textbf{Step (4)}, we use Theorem \ref{thm:G-CG-Net Lipschitz} and Corollary \ref{corollary:covering number bound for multiple Lipschitz bounds} to bound the covering numbers from Dudley's inequality by covering numbers of the G-CG-Net parameter spaces. \textbf{Step (5)}, we use Lemma \ref{lemma:covering number bound on unit subset} to bound the covering numbers of the G-CG-Net parameter spaces. \textbf{Step (6)}, with the Step (5) bounds, we bound Dudley's inequality by evaluable integrals where evaluation, simplification, and using Theorem~\ref{thm:GEB by Rademacher complexity} produces the desired GEB for G-CG-Net.
\begin{proof}[Proof of Theorem \ref{thm:GEB G-CG-Net}] \textbf{Step (1).}
    Let $\overline{Y} \coloneqq [\overline{\bm{y}}_1, \ldots, \overline{\bm{y}}_{N_s}]\in\mathbb{R}^{m\times N_s}$ and
\begin{align*}
    \mathcal{M}_{{\textnormal{CG}}}^{(1)} &\coloneqq \{M_{\widehat{\bm{c}}} = [\widehat{\bm{c}}(\overline{\bm{y}}_1), \ldots, \widehat{\bm{c}}(\overline{\bm{y}}_{N_s})]: \widehat{\bm{c}}\in\mathcal{H}_{\textnormal{CG}}^{(1)}\} \\
    &= \scalebox{.99}{${\displaystyle\left\{\widehat{\bm{c}}\left(\overline{Y}; \left\{P,\bm{\theta}_{k,d}^{(j)}\right\}\middlescript{\substack{j \in\mathbb{N}[J] \\ k \in\mathbb{N}[K] \\ d \in\mathbb{N}[D]}}\right): P \in\mathcal{P}_{\textnormal{full}}, \bm{\theta}_{k,d}^{(j)}\in\Omega_{d}\right\}}$}
\end{align*}
where $ \widehat{\bm{c}}\left(\overline{Y}; \bm{\Theta}\right) \coloneqq [\widehat{\bm{c}}\left(\overline{\bm{y}}_1; \bm{\Theta}\right), \ldots, \widehat{\bm{c}}\left(\overline{\bm{y}}_{N_s}; \bm{\Theta}\right)] \in\mathbb{R}^{n\times N_s}.$ Now, define the Rademacher process $(X_{M_{\widehat{\bm{c}}}})_{{M_{\widehat{\bm{c}}}}\in \mathcal{M}_{{\textnormal{CG}}}^{(1)}}$ as
\begin{align}
    X_{M_{\widehat{\bm{c}}}} \coloneqq \sum_{i = 1}^{N_s}\sum_{k = 1}^n \gamma_{ik} [M_{\widehat{\bm{c}}}]_{ki} = \sum_{i = 1}^{N_s}\sum_{k = 1}^n \gamma_{ik} \widehat{c}_k(\overline{\bm{y}}_i) \label{eqn:NN random process}
\end{align}
for Rademacher variables $\Gamma = [\gamma_{ik}]_{i \in\mathbb{N}[N_s]}^{k\in\mathbb{N}[n]}$. 

\textbf{Step (2).} Using Lemma \ref{lemma:contraction lemma} with $\mathcal{H} = \mathcal{H}_{\textnormal{CG}}^{(1)}$ and each $g_i = L$, for $L$ a $\tau$-Lipschitz loss function of G-CG-Net, e.g. SSIM loss or mean-absolute error, satisfying Assumption \ref{assumption:loss function}, we have
\begin{align}
    \mathcal{R}_{\mathcal{S}}(L\circ\mathcal{H}_{\textnormal{CG}}^{(1)}) &\leq \sqrt{2}\tau \mathbb{E}_{\Gamma}\left[\sup_{\widehat{\bm{c}}\in\mathcal{H}_{\textnormal{CG}}^{(1)}} \frac{1}{N_s}\sum_{i = 1}^{N_s}\sum_{k = 1}^n \gamma_{ik} \widehat{c}_k(\overline{\bm{y}}_i)\right] \nonumber \\
    &= \frac{\sqrt{2}\tau}{N_s} \mathbb{E}_{\Gamma}\bigg(\sup_{\widehat{\bm{c}}\in\mathcal{H}_{\textnormal{CG}}^{(1)}} X_{M_{\widehat{\bm{c}}}}\bigg). \label{eqn:Rademacher complexity bound 1}
\end{align}

\textbf{Step (3).} By equation (\ref{eqn:pseudometric}), note that
\begin{align*}
    \widetilde{d}(X_{M_{\widehat{\bm{c}}_1}}, X_{M_{\widehat{\bm{c}}_2}})^2 &= \sum_{i = 1}^{N_s}\sum_{k = 1}^n ([{M_{\widehat{\bm{c}}_1}}]_{ki} - [M_{\widehat{\bm{c}}_2}]_{ki})^2 \\
    &= \norm{M_{\widehat{\bm{c}}_1}-M_{\widehat{\bm{c}}_2}}_F^2
\end{align*}
for $\norm{\cdot}_F$ the Frobenius norm. Observe for any $M_{\widehat{\bm{c}}}\in \mathcal{{M}}_{\textnormal{CG}}^{(1)}$
\begin{align*}
    \norm{M_{\widehat{\bm{c}}}}_F = \norm{\widehat{\bm{c}}(\overline{Y};\bm{\Theta})}_F = \sqrt{\sum_{i = 1}^{N_s} \norm{\widehat{\bm{c}}(\overline{\bm{y}}_i;\bm{\Theta})}_2^2} \leq \sqrt{N_s} c_{\max}
\end{align*}
where we used that any output from G-CG-Net is bounded, in Euclidean norm, by $c_{\max}.$ Hence, by equation (\ref{eqn:expectation of Rademacher process squared})
\begin{align*}
    \Delta(\mathcal{M}_{\textnormal{CG}}^{(1)}) = \scalebox{1}{$\sup_{\widehat{\bm{c}}\in \mathcal{H}_{\textnormal{CG}}^{(1)}}$} \norm{M_{\widehat{\bm{c}}}}_F \leq \sqrt{N_s} c_{\max}.
\end{align*}
Therefore, using Dudley's Inequality in Lemma~\ref{lemma:dudley's inequality} and (\ref{eqn:Rademacher complexity bound 1})
\begin{align}
   \scalebox{1}{{$\displaystyle \mathcal{R}_{\mathcal{S}}(L\circ \mathcal{H}_{\textnormal{CG}}^{(1)}) \leq \frac{8\tau}{N_s} \int_0^{\frac{\sqrt{N_s}c_{\max}}{2}} \sqrt{\ln\left(\mathcal{N}(\mathcal{M}_{\textnormal{CG}}^{(1)},\norm{\cdot}_F, \epsilon)\right)} d\epsilon $}}. \label{eqn:rademacher complexity bound 3}
\end{align}

\textbf{Step (4).} By Theorem \ref{thm:G-CG-Net Lipschitz} for any $i\in\mathbb{N}[N_s]$
    \begin{align*}
        &\norm{\widehat{\bm{c}}(\overline{\bm{y}}_i;\bm{\Theta}) - \widehat{\bm{c}}(\overline{\bm{y}}_i;\widetilde{\bm{\Theta}})}_2 \\
        &\leq \kappa \norm{P_u - \widetilde{P}_u}_2 + \sum_{k = 1}^K\sum_{j = 1}^J\sum_{d = 1}^D \kappa_{k,d}^{(j)} \norm{\bm{\theta}_{k, d}^{(j)} - \widetilde{\bm{\theta}}_{k,d}^{(j)}}_{(d)}.
    \end{align*}
    As $\kappa$ and $\kappa_{k,d}^{(j)}$, given respectively in (\ref{eqn:kappa}) and (\ref{eqn:kappa k,d,j}), are independent of $\overline{\bm{y}}_i$ then
    \begin{align*}
        \norm{\widehat{\bm{c}}(\overline{Y}; \bm{\Theta}) - \widehat{\bm{c}}(\overline{Y};\widetilde{\bm{\Theta}})}_F 
        &\leq \sqrt{N_s}\norm{\widehat{\bm{c}}(\overline{\bm{y}}_i;\bm{\Theta}) - \widehat{\bm{c}}(\overline{\bm{y}}_i;\widetilde{\bm{\Theta}})}_2
    \end{align*}
    for any $i \in\mathbb{N}[N_s]$. Thus
    \begin{align*}
        &\norm{\widehat{\bm{c}}(\overline{Y};\bm{\Theta}) - \widehat{\bm{c}}(\overline{Y};\widetilde{\bm{\Theta}})}_F \leq \sqrt{N_s}\kappa \norm{P_u - \widetilde{P}_u}_2 + \\
        & \sqrt{N_s}\sum_{k = 1}^K\sum_{j = 1}^J\sum_{d = 1}^D \kappa_{k,d}^{(j)} \norm{\bm{\theta}_{k, d}^{(j)} - \widetilde{\bm{\theta}}_{k,d}^{(j)}}_{(d)}.
    \end{align*}
    Hence, from Corollary \ref{corollary:covering number bound for multiple Lipschitz bounds}
    \begin{align}
        &\ln\left(\mathcal{N}\left(\mathcal{M}_{\textnormal{CG}}^{(1)}, \norm{\cdot}_{F}, \epsilon\right)\right) \nonumber \\
        &\leq \ln\left(\mathcal{N}\left(\mathcal{P}, \norm{\cdot}_2, \frac{\epsilon}{\sqrt{N_s}\kappa (KJD+1)}\right)\right) \nonumber \\
        &\resizebox{\columnwidth}{!}{${\displaystyle+\sum_{k = 1}^K\sum_{j = 1}^J\sum_{d = 1}^D \ln\left(\mathcal{N}\left(\Omega_{d}, \norm{\cdot}_{(d)}, \frac{\epsilon}{\sqrt{N_s}\kappa_{k,d}^{(j)} (KJD+1)}\right)\right).}$} \label{eqn:log covering number bound 1}
    \end{align}

    \textbf{Step (5).} As $\mathcal{P}$ contains symmetric $n\times n$ matrices, then for $\mathcal{P} = \mathcal{P}_{\textnormal{full}}$ only $n(n+1)/2$ entries are required to uniquely define a matrix and $\dim(\mathcal{P}_{\textnormal{full}}) = n(n+1)/2$. Similarly, $\dim(\mathcal{P}_{\textnormal{tri}}) = 2n-1$, $\dim(\mathcal{P}_{\textnormal{diag}}) = n$, and $\dim(\mathcal{P}_{\textnormal{const}}) = 1$. Furthermore, as $\mathcal{P}/p_{\max} = \{P/p_{\max} : P\in\mathcal{P}\}$ is contained in the unit $\norm{\cdot}_2$ ball, then, using Lemma \ref{lemma:covering number bound on unit subset}, observe
    \begin{align*}
        \mathcal{N}\left(\mathcal{P}, \norm{\cdot}_2, \epsilon\right) &= \mathcal{N}\left(\mathcal{P}/p_{\max}, \norm{\cdot}_2, \epsilon/p_{\max}\right) \\
        &\leq \left(1+2p_{\max}/\epsilon\right)^{\dim(\mathcal{P})}.
    \end{align*}
    Similarly, as $\Omega_d/\omega_d$ is contained in the unit $\norm{\cdot}_{(d)}$ ball, observe
    \begin{align*}
        \mathcal{N}\left(\Omega_{d}, \norm{\cdot}_{(d)}, \epsilon\right) \leq \left(1+2\omega_{d}/\epsilon\right)^{\alpha_{d}}.
    \end{align*}
    Combining these two observations with (\ref{eqn:log covering number bound 1})
    gives
    \begin{align}
        &\ln\left(\mathcal{N}\left(\mathcal{M}_{\textnormal{CG}}^{(1)}, \norm{\cdot}_{F}, \epsilon\right)\right) \nonumber \\
        &\leq \dim(\mathcal{P})\ln\left(1+\frac{2p_{\max}\sqrt{N_s}\kappa (KJD+1)}{\epsilon}\right) \nonumber \\
        &+\sum_{k = 1}^K\sum_{j = 1}^J\sum_{d = 1}^D \alpha_{d} \ln\left(1+\frac{2\omega_{d}\sqrt{N_s}\kappa_{k,d}^{(j)} (KJD+1)}{\epsilon}\right). \label{eqn:log covering number bound 2}
    \end{align}

    \textbf{Step (6).} For any $\nu, \beta > 0$, note that from~\cite{NNfromstatistical}
    \begin{align}
        \scalebox{1}{$\int_0^\beta$} \sqrt{\ln\left(1+\nu/\epsilon\right)}d\epsilon \leq \beta \sqrt{\ln(e(1+\nu/\beta))}. \label{eqn:integral of log bound}
    \end{align}
    Combining (\ref{eqn:log covering number bound 2}), (\ref{eqn:integral of log bound}), and using subadditivity of square roots
    \begin{align}
        &\int_0^{\beta} \sqrt{\ln\left(\mathcal{N}\left(\mathcal{M}_{\textnormal{CG}}^{(1)}, \norm{\cdot}_{F}, \epsilon\right)\right)} d\epsilon \nonumber \\
        &\scalebox{.9}{${\displaystyle \leq \sqrt{\dim(\mathcal{P})} \int_0^{\beta} \sqrt{\ln\left(1+\frac{2p_{\max}\sqrt{N_s}\kappa (KJD+1)}{\epsilon}\right)}d\epsilon}$} \nonumber \\
        &\scalebox{.85}{${\displaystyle+ \sum_{k = 1}^K\sum_{j = 1}^J\sum_{d = 1}^D \sqrt{\alpha_{d}}\int_0^{\beta} \sqrt{\ln\left(1+\frac{2\omega_{d}\sqrt{N_s}\kappa_{k,d}^{(j)} (KJD+1)}{\epsilon}\right)}d\epsilon}$} \nonumber \\
        &\scalebox{.9}{${\displaystyle \leq \sqrt{\dim(\mathcal{P})} \beta \sqrt{\ln\left(e\left(1+\frac{2p_{\max}\sqrt{N_s}\kappa (KJD+1)}{\beta}\right)\right)}}$} \nonumber \\
        &\scalebox{.85}{${\displaystyle+ \sum_{k = 1}^K\sum_{j = 1}^J\sum_{d = 1}^D \sqrt{\alpha_{d}}\beta \sqrt{\ln\left(e\left(1+\frac{2\omega_{d}\sqrt{N_s}\kappa_{k,d}^{(j)} (KJD+1)}{\beta}\right)\right)}}$.} \label{eqn:integral of covering number bound}
    \end{align}
    
    Combining (\ref{eqn:integral of covering number bound}) and (\ref{eqn:rademacher complexity bound 3}), for $\beta = \sqrt{N_s}c_{\max}/2,$ with Theorem \ref{thm:GEB by Rademacher complexity} produces the desired generalization error bound.
\end{proof}

\section{Appendix: Proof of Theorem \ref{thm:GEB CG-Net}} \label{apndx:CG-Net GEB proof}

First, we require a Lipschitz condition on the gradient of the data fidelity term in (\ref{eqn:cost function}) that depends on Proposition \ref{prop:u bound}.
\begin{lemma} \label{lemma:Lipschitz bound on Au(Auz-y)}
    For every $\bm{z}_i\in\mathbb{R}^n$ satisfying $\norm{\bm{z}_i}_\infty \leq z_\infty$ and $\bm{u}_i = \mathcal{T}_{\bm{y}}(\bm{z}_i; P_i)$ where $P_i\in\mathcal{P}$, it holds that
    \begin{align*}
        &\norm{A_{\bm{u}_1}^T(A_{\bm{u}_1}\bm{z}_1-\bm{y}) -  A_{\bm{u}_2}^T(A_{\bm{u}_2}\bm{z}_2-\bm{y})}_2 \nonumber\\
        &\leq (z_{\infty} p_{\max}\norm{\bm{y}}_2 \norm{A}_2 \norm{A}_{\infty})^2 \norm{\bm{z}_1 - \bm{z}_2}_2 \nonumber\\
        &\resizebox{\columnwidth}{!}{{$\displaystyle + \norm{\bm{y}}_2\norm{A}_2\left(1+z_{\infty}^2p_{\max} \norm{A}_2(\norm{A}_2+\norm{A}_{\infty})\right)\norm{\bm{u}_1 - \bm{u}_2}_2$.}}
    \end{align*}
\end{lemma}
\begin{proof}
Observe
    \begin{align}
        &\norm{A_{\bm{u}_1}^T(A_{\bm{u}_1}\bm{z}_1-\bm{y}) -  A_{\bm{u}_2}^T(A_{\bm{u}_2}\bm{z}_2-\bm{y})}_2 \nonumber\\
        &\leq \norm{A_{\bm{u}_1}^TA_{\bm{u}_1}\bm{z}_1 - A_{\bm{u}_2}^TA_{\bm{u}_2}\bm{z}_2}_2 + \norm{(A_{\bm{u}_1}^T - A_{\bm{u}_2}^T)\bm{y}}_2 \nonumber\\
        &\leq \norm{A_{\bm{u}_1}^TA_{\bm{u}_1}\bm{z}_1 - A_{\bm{u}_2}^TA_{\bm{u}_2}\bm{z}_2}_2 + \norm{A}_2\norm{\bm{y}}_2 \norm{\bm{u}_1 - \bm{u}_2}_2. \label{eqn:Lipschitz bound on Au(Auz-y) 1}
    \end{align}
By Proposition \ref{prop:u bound} and the triangle inequality note that
    \begin{align}
        &\norm{A_{\bm{u}_1}^TA_{\bm{u}_1}\bm{z}_1 - A_{\bm{u}_2}^TA_{\bm{u}_2}\bm{z}_2}_2 \nonumber \\
        &= \left\lVert A_{\bm{u}_1}^TA_{\bm{u}_1}\bm{z}_1 - A_{\bm{u}_1}^TA_{\bm{u}_2}\bm{z}_1+A_{\bm{u}_1}^TA_{\bm{u}_2}\bm{z}_1-A_{\bm{u}_2}^TA_{\bm{u}_2}\bm{z}_1 \right. \nonumber \\
        &\hspace{.65cm}\left.+A_{\bm{u}_2}^TA_{\bm{u}_2}\bm{z}_1-A_{\bm{u}_2}^TA_{\bm{u}_2}\bm{z}_2\right\rVert_2 \nonumber \\
        &\leq z_{\infty}^2p_{\max} \norm{A}_2^2\, \norm{\bm{y}}_2 (\norm{A}_2+\norm{A}_{\infty})\norm{\bm{u}_1 - \bm{u}_2}_2 \nonumber \\
        &\hspace{.65cm} + (z_{\infty} p_{\max} \norm{A}_2 \norm{A}_{\infty} \norm{\bm{y}}_2)^2 \norm{\bm{z}_1 - \bm{z}_2}_2. \label{eqn:Lipschitz bound on AuAuz}
    \end{align}
    Combining (\ref{eqn:Lipschitz bound on Au(Auz-y) 1}) and (\ref{eqn:Lipschitz bound on AuAuz}) produces the desired result.
\end{proof}

As an overview, proving Theorem \ref{thm:GEB CG-Net} consists of the following steps: First, we show Assumption \ref{assumption:loss function} holds for the SSIM loss function used in CG-Net. Second, we invoke Lemma \ref{lemma:Lipschitz bound on Au(Auz-y)} to show that the Lipschitz condition of Assumption \ref{assumption:scale update method} holds for each CG-Net scale-variable-descent update in (\ref{eqn:CG-Net scale variable descent update}). Finally, we apply Theorem \ref{thm:GEB G-CG-Net} to produce Theorem \ref{thm:GEB CG-Net}.

\begin{proof}[Proof of Theorem \ref{thm:GEB CG-Net}]
    As SSIM returns a value in $[-1,1]$ then the SSIM loss function is bounded by $2$. From~\cite{NN_loss}, SSIM is a differentiable function and thus continuous. As the CG-Net outputs are bounded, in the $\norm{\cdot}_2$ ball of radius $c_{\max}$, then the gradient of the SSIM loss function is bounded. Hence, by the mean value theorem~\cite{rudin1953principles}, there exists a Lipschitz constant for the SSIM loss function, on the $\norm{\cdot}_2$ ball of radius $c_{\max}$, which we denote by $\tau$. Therefore, Assumption \ref{assumption:loss function} holds.

    Next, for $i \in \{1, 2\},$ let $\bm{z}_i\in\mathbb{R}^n$ satisfy $\norm{\bm{z}_i}_\infty\leq z_\infty$ and $\bm{u}_i = \mathcal{T}_{\overline{\bm{y}}_p}(\bm{z}_i; P_i)$ for some $P_i\in\mathcal{P}_{\textnormal{const}}$ and $p\in\mathbb{N}[N_s]$. As $\mathcal{P}_{a,b}$ is 1-Lipschitz, then the CG-Net scale-variable-update method, $g_k^{(j)}$, in (\ref{eqn:CG-Net scale variable descent update}) satisfies
    \begin{align}
        &\norm{g_k^{(j)}(\bm{z}_1,\bm{u}_1; B_k^{(j)}, \mu_k^{(j)}) - g_k^{(j)}(\bm{z}_2,\bm{u}_2; \widetilde{B}_k^{(j)}, \widetilde{\mu}_k^{(j)})}_2 \nonumber \\
        &= \left\lVert\mathcal{P}_{a,b}(\bm{z}_1 - B_{k}^{(j)} \rho_{\xi}(\nabla_{\bm{z}} F(\bm{u}_1,\bm{z}_1; \mu_k^{(j)}))) \right. \nonumber \\ 
        &\hspace{.65cm} \left.- \mathcal{P}_{a,b}(\bm{z}_2 - \widetilde{B}_{k}^{(j)} \rho_{\xi}(\nabla_{\bm{z}} F(\bm{u}_2,\bm{z}_2; \widetilde{\mu}_k^{(j)})))\right\rVert_2 \nonumber \\ 
        &\leq \norm{\bm{z}_1 - B_{k}^{(j)} \rho_{\xi}(\nabla_{\bm{z}} F(\bm{u}_1,\bm{z}_1; \mu_k^{(j)})) \nonumber \\ 
        &\hspace{.65cm} - \bm{z}_2 + \widetilde{B}_{k}^{(j)} \rho_{\xi}(\nabla_{\bm{z}} F(\bm{u}_2,\bm{z}_2; \widetilde{\mu}_k^{(j)}))}_2 \nonumber \\ 
        &\leq \norm{\bm{z}_1 - \bm{z}_2}_2 + \left\lVert\widetilde{B}_{k}^{(j)} \rho_{\xi}(\nabla_{\bm{z}} F(\bm{u}_2,\bm{z}_2; \widetilde{\mu}_k^{(j)})) \right.\nonumber  \\ 
        &\hspace{2.65cm} \left. - B_{k}^{(j)} \rho_{\xi}(\nabla_{\bm{z}} F(\bm{u}_1,\bm{z}_1; \mu_k^{(j)}))\right\rVert_2 \label{eqn:CG-Net scale variable update bound 1}
    \end{align}
    where in the final line we used the triangle inequality. First, as $\rho_{\xi}$ is 1-Lipschitz and bounded by $\xi$, note that
    \begin{align}
        &\resizebox{\columnwidth}{!}{${\displaystyle \norm{\widetilde{B}_{k}^{(j)} \rho_{\xi}(\nabla_{\bm{z}} F(\bm{u}_2,\bm{z}_2; \widetilde{\mu}_k^{(j)}))  - B_{k}^{(j)} \rho_{\xi}(\nabla_{\bm{z}} F(\bm{u}_1,\bm{z}_1; \mu_k^{(j)}))}_2}$} \nonumber \\
        &\resizebox{\columnwidth}{!}{${\displaystyle= \left\lVert\widetilde{B}_{k}^{(j)} \rho_{\xi}(\nabla_{\bm{z}} F(\bm{u}_2,\bm{z}_2; \widetilde{\mu}_k^{(j)})) - \widetilde{B}_{k}^{(j)} \rho_{\xi}(\nabla_{\bm{z}} F(\bm{u}_1,\bm{z}_1; \mu_k^{(j)})) \right.}$}  \nonumber \\
        &\resizebox{\columnwidth}{!}{${\displaystyle \left.+\widetilde{B}_{k}^{(j)} \rho_{\xi}(\nabla_{\bm{z}} F(\bm{u}_1,\bm{z}_1; \mu_k^{(j)}))- B_{k}^{(j)} \rho_{\xi}(\nabla_{\bm{z}} F(\bm{u}_1,\bm{z}_1; \mu_k^{(j)}))\right\rVert_2 }$} \nonumber \\
        &\leq p_{\max} \norm{  \rho_{\xi}(\nabla_{\bm{z}} F(\bm{u}_2,\bm{z}_2; \widetilde{\mu}_k^{(j)})) - \rho_{\xi}(\nabla_{\bm{z}} F(\bm{u}_1,\bm{z}_1; \mu_k^{(j)}))}_2 \nonumber \\ 
        &\hspace{.65cm} + \norm{\rho_{\xi}(\nabla_{\bm{z}} F(\bm{u}_1,\bm{z}_1; \mu_k^{(j)}))}_2\, \norm{\widetilde{B}_{k}^{(j)} - B_{k}^{(j)}}_2 \nonumber \\ 
        &\leq p_{\max} \norm{\nabla_{\bm{z}} F(\bm{u}_2,\bm{z}_2; \widetilde{\mu}_k^{(j)}) - \nabla_{\bm{z}} F(\bm{u}_1,\bm{z}_1; \mu_k^{(j)})}_2 \nonumber \\ 
        &\hspace{.65cm} + \xi \, \norm{\widetilde{B}_{k}^{(j)} - B_{k}^{(j)}}_2. \label{eqn:CG-Net scale variable update bound 2} 
    \end{align}
    Second, note that
    \begin{align}
        & \norm{\nabla_{\bm{z}} F(\bm{u}_2,\bm{z}_2; \widetilde{\mu}_k^{(j)}) - \nabla_{\bm{z}} F(\bm{u}_1,\bm{z}_1; \mu_k^{(j)})}_2 \nonumber  \\
        &\leq \norm{A_{\bm{u}_1}^T(A_{\bm{u}_1}\bm{z}_1-\overline{\bm{y}}_p) -  A_{\bm{u}_2}^T(A_{\bm{u}_2}\bm{z}_2-\overline{\bm{y}}_p)}_2 \nonumber \\
        & +  \norm{\mu_k^{(j)}  [h^{-1}]'(\bm{z}_1)\odot h^{-1}(\bm{z}_1) - \widetilde{\mu}_k^{(j)}  [h^{-1}]'(\bm{z}_2)\odot h^{-1}(\bm{z}_2)}_2 \nonumber \\
        &= \norm{A_{\bm{u}_1}^T(A_{\bm{u}_1}\bm{z}_1-\overline{\bm{y}}_p) -  A_{\bm{u}_2}^T(A_{\bm{u}_2}\bm{z}_2-\overline{\bm{y}}_p)}_2 + \nonumber \\
        & \left\lVert\mu_k^{(j)}  [h^{-1}]'(\bm{z}_1)\odot h^{-1}(\bm{z}_1) -\widetilde{\mu}_k^{(j)}  [h^{-1}]'(\bm{z}_1)\odot h^{-1}(\bm{z}_1) \right.\nonumber \\
        &\hspace{.25cm} \left.+\widetilde{\mu}_k^{(j)}  [h^{-1}]'(\bm{z}_1)\odot h^{-1}(\bm{z}_1) - \widetilde{\mu}_k^{(j)}  [h^{-1}]'(\bm{z}_2)\odot h^{-1}(\bm{z}_2)\right\rVert_2 \nonumber \\
        & \leq \norm{A_{\bm{u}_1}^T(A_{\bm{u}_1}\bm{z}_1-\overline{\bm{y}}_p) -  A_{\bm{u}_2}^T(A_{\bm{u}_2}\bm{z}_2-\overline{\bm{y}}_p)}_2 \nonumber \\
        &\hspace{.5cm} +  h_{\max} |\mu_k^{(j)}  -\widetilde{\mu}_k^{(j)}| + \mu \tau_h \norm{\bm{z}_1 - \bm{z}_2}_2. \label{eqn:gradient bound 1} 
    \end{align}

Combining Lemma \ref{lemma:Lipschitz bound on Au(Auz-y)} with (\ref{eqn:gradient bound 1}), (\ref{eqn:CG-Net scale variable update bound 2}), (\ref{eqn:CG-Net scale variable update bound 1}) and the fact that $\norm{\overline{\bm{y}}_p}_2\leq y_{\max}$, then Assumption \ref{assumption:scale update method} holds specifically with $r_{k,1}^{(j-1)} = r_1, r_{k,2}^{(j-1)} = r_2$ and
    \begin{align*}
        \left(\bm{\theta}_{k,d}^{(j)}, r_{k,d,3}^{(j-1)}, \norm{\cdot}_{(d)}\right) = \begin{cases}
            \left(B_{k}^{(j)}, \xi, \norm{\cdot}_2\right) & d = 1 \\
            \left(\mu_k^{(j)}, p_{\max}h_{\max}, |\cdot|\right) & d = 2.
        \end{cases}
    \end{align*}
    Hence, by Proposition \ref{prop:scale variable lipschitz 1}, $ \widehat{r}_{k,1}^{(J)} = \prod_{j = 1}^{J} r_{k,1}^{(j-1)} = r_1^J,$
\begin{align*}
    \widehat{r}_{k,2}^{(J)} &= \sum_{j = 1}^{J} r_{k,2}^{(j-1)}\prod_{\ell = j}^{J-1} r_{k,1}^{(\ell)} = r_2 \sum_{j =1}^J r_1^{J-j} = r_2\frac{1-r_1^J}{1-r_1},
\end{align*}
and similarly $\widehat{r}_{k,d,3}^{(j,J)} = r_{k,d,3}^{(j-1)}\prod_{\ell = j}^{J-1} r_{k,1}^{(\ell)} = r_{k,d,3}^{(j-1)} r_1^{J-j}$ is as given in (\ref{eqn:r_3 CG-Net}). Therefore, applying Theorem~\ref{thm:GEB G-CG-Net} produces the GEB of CG-Net in Theorem~\ref{thm:GEB CG-Net}.
\end{proof}

\section{Appendix: Proof of Theorem \ref{thm:GEB DR-CG-Net}} \label{apndx:DR-CG-Net GEB proof}

We first require a bounded and Lipschitz property for fully-connected networks. Let $\mathcal{G}_t^{(i)}(\bm{x}) \coloneqq W_t^{(i)}\bm{x}$, for $W_t^{(i)}\in\mathbb{R}^{d_{t+1}\times d_t}$, denote a fully-connected layer. For componentwise activation function $\sigma$, define a fully-connected network $\mathcal{G}^{(i, T)}$ as
\begin{align}
    \mathcal{G}^{(i,T)}(\bm{x}) = \mathcal{G}_T^{(i)} \circ \sigma \circ \mathcal{G}_{T-1}^{(i)} \circ \cdots  \circ \sigma \circ \mathcal{G}_1^{(i)}(\bm{x}). \label{eqn:fully connected network}
\end{align}
\begin{lemma}\label{lemma:bnd fully connected networks}
    Let $\sigma$ be a componentwise activation function satisfying $\norm{\sigma(\bm{x})}_2\leq \norm{\bm{x}}_2$ and $\mathcal{G}^{(i,T)}$ be given as in (\ref{eqn:fully connected network}). If $\norm{W_t^{(i)}}_2\leq \varpi_t$ for all $t\in\mathbb{N}[T]$, then 
    \begin{align*}
        \norm{\sigma(\mathcal{G}^{(i,T)}(\bm{x}))}_2 \leq \prod_{t = 1}^T \varpi_t \norm{\bm{x}}_2.
    \end{align*}
\end{lemma}
\begin{proof}
    We use induction on $T$. The base case $T = 1$ holds trivially. Assume the induction hypothesis holds for fixed $T$ where $T > 1$. Observe
    \begin{align*}
        \norm{\sigma(\mathcal{G}^{(i,T+1)}(\bm{x}))}_2 
        &= \norm{\sigma(W_{T+1}^{(i)} \sigma(\mathcal{G}^{(i,T)}(\bm{x})))}_2 \\
        &\leq \varpi_{T+1} \norm{\sigma(\mathcal{G}^{(i,T)}(\bm{x}))}_2 \leq \prod_{t = 1}^{T+1} \varpi_t \norm{\bm{x}}_2. \qedhere
    \end{align*}
\end{proof}
Now, we show that fully-connected networks are Lipschitz.
\begin{lemma}\label{lemma:lipschitz of fully-connected network}
In addition to the assumptions of Lemma \ref{lemma:bnd fully connected networks}, assume $\sigma$ is $\tau$-Lipschitz. Then
\begin{multline*}
\norm{\mathcal{G}^{(1, T)}(\bm{x}_1) - \mathcal{G}^{(2, T)}(\bm{x}_2)}_2 \leq \tau^{T-1} \prod_{t = 1}^T \varpi_t \norm{\bm{x}_1 - \bm{x}_2}_2 \\
 + \sum_{t = 1}^T \bigg(\tau^{T - t}\scalebox{1}{$\prod_{\substack{t' = 1 \\ t'\neq t}}^{T}$}\varpi_{t'} \norm{\bm{x}_1}_2\bigg) \norm{W_t^{(1)} - W_t^{(2)}}_2.
    \end{multline*}
\end{lemma} 
\begin{proof}
   Observe for any $t$
   \begin{align*}
       &\norm{\mathcal{G}_t^{(1)}(\bm{x}_1) - \mathcal{G}_t^{(2)}(\bm{x}_2)}_2 \\
       &= \norm{W_t^{(1)}\bm{x}_1 - W_t^{(2)}\bm{x}_2}_2 \\
       &=  \norm{W_t^{(1)}\bm{x}_1-W_t^{(2)}\bm{x}_1 +W_t^{(2)}\bm{x}_1 - W_t^{(2)}\bm{x}_2}_2 \\
       &\leq \varpi_t \norm{\bm{x}_1 - \bm{x}_2}_2 + \norm{\bm{x_1}}_2\, \norm{W_t^{(1)} - W_t^{(2)}}_2.
   \end{align*}
   Therefore
   \begin{multline*}
       \norm{\mathcal{G}^{(1,t)}(\bm{x}_1) - \mathcal{G}^{(2,t)}(\bm{x}_2)}_2 \\
       \leq \tau \varpi_t \norm{\mathcal{G}^{(1,t-1)}(\bm{x}_1) - \mathcal{G}^{(2, t-1)}(\bm{x}_2)}_2 \\
       + \norm{\sigma(\mathcal{G}^{(1,t-1)}(\bm{x}_1))}_2\, \norm{W_t^{(1)} - W_t^{(2)}}_2.
   \end{multline*}
   Using Lemma \ref{lemma:bnd fully connected networks} and induction on $T$ similar to that of Proposition \ref{prop:scale variable lipschitz 1} produces the desired result.
\end{proof}

As an overview, proving Theorem \ref{thm:GEB DR-CG-Net} consists of the following steps: First, we show Assumption \ref{assumption:loss function} holds for the mean-absolute error loss function used in DR-CG-Net. Second, we use invoke Lemma \ref{lemma:Lipschitz bound on Au(Auz-y)} and Lemma \ref{lemma:lipschitz of fully-connected network} to show that the Lipschitz condition of Assumption \ref{assumption:scale update method} holds for each DR-CG-Net scale-variable-descent update in (\ref{eqn:PGD scale variable update}). Finally, we apply Theorem \ref{thm:GEB G-CG-Net} to produce Theorem \ref{thm:GEB DR-CG-Net}.

\begin{proof}[Proof of Theorem \ref{thm:GEB DR-CG-Net}]
    As DR-CG-Net employs the mean absolute loss function, $L(\bm{x}_1,\bm{x}_2) = \frac{1}{n}\norm{\bm{x}_1-\bm{x}_2}_1$, for any $\bm{x}_1, \bm{x}_2, \bm{x}\in\mathbb{R}^n$ by H\"{o}lder's and the reverse triangle inequality
\begin{align*}
   &|L(\bm{x}_1,\bm{x}) - L(\bm{x}_2, \bm{x})| \\
   &\leq  L(\bm{x}_1,\bm{x}_2) = \frac{1}{n}\norm{\bm{x}_1 - \bm{x}_2}_1 \leq \frac{1}{\sqrt{n}} \norm{\bm{x}_1 - \bm{x}_2}_2 \leq \frac{c_{\max}}{\sqrt{n}}.
\end{align*}
Hence, Assumption \ref{assumption:loss function} holds for $c = c_{\max}/\sqrt{n}$ and $\tau = 1/\sqrt{n}$.

Next, for $i \in \{1, 2\},$ let $\bm{z}_i\in\mathbb{R}^n$ satisfy $\norm{\bm{z}_i}_\infty\leq z_\infty$ and $\bm{u}_i = \mathcal{T}_{\overline{\bm{y}}_p}(\bm{z}_i; P_i)$ for some $P_i\in\mathcal{P}_{\textnormal{tri}}$ and $p\in\mathbb{N}[N_s]$. Let $\mathcal{V}_k^{(j)}$ and $\widetilde{\mathcal{V}}_k^{(j)}$ be convolutional subnetworks of DR-CG-Net that are parameterized by $\{W_{k,\ell}^{(j)}\}_{\ell\in\mathbb{N}[L_c]}$ and $\{\widetilde{W}_{k,\ell}^{(j)}\}_{\ell\in\mathbb{N}[L_c]}$, respectively. The DR-CG-Net scale-variable-update method, $g_k^{(j)}$, in (\ref{eqn:PGD scale variable update}) satisfies
\begin{align}
 &\left\lVert g_k^{(j)}(\bm{z}_1,\bm{u}_1; \{\delta_k^{(j)}, W_{k,\ell}^{(j)}\}_{\ell\in\mathbb{N}[L_c]}) \right. \nonumber \\
 &\hspace{.25cm} \left.- g_k^{(j)}(\bm{z}_2,\bm{u}_2; \{\widetilde{\delta}_k^{(j)}, \widetilde{W}_{k,\ell}^{(j)}\}_{\ell\in\mathbb{N}[L_c]})\right\rVert_2 \nonumber \\
 &= \left\lVert v_k^{(j)}(\bm{z}_1,\bm{u}_1; \delta_{k}^{(j)}) + \mathcal{V}_k^{(j)}\left(\bm{z}_1\right) \right. \nonumber \\
 &\hspace{.5cm} \left.- v_k^{(j)}(\bm{z}_2,\bm{u}_2; \widetilde{\delta}_{k}^{(j)}) - \widetilde{\mathcal{V}}_k^{(j)}\left(\bm{z}_2\right)\right\rVert_2 \nonumber \\
 &\leq \norm{v_k^{(j)}(\bm{z}_1,\bm{u}_1; \delta_{k}^{(j)}) - v_k^{(j)}(\bm{z}_2,\bm{u}_2; \widetilde{\delta}_{k}^{(j)})}_2 \nonumber \\
 &\hspace{.5cm} + \norm{\mathcal{V}_k^{(j)}\left(\bm{z}_1\right) - \widetilde{\mathcal{V}}_k^{(j)}\left(\bm{z}_2\right)}_2. \label{eqn:scale mapping module bound 1 DR-CG-Net}
\end{align}

    Define $\bm{d}_i = A_{\bm{u}_i}^T(A_{\bm{u}_i}\bm{z}_i-\overline{\bm{y}}_p)$ for $i = 1, 2$. Using (\ref{eqn:data fidelity gradient update}) note
    \begin{multline}
        \norm{v_k^{(j)}(\bm{z}_1,\bm{u}_1; \delta_{k}^{(j)}) - v_k^{(j)}(\bm{z}_2,\bm{u}_2; \widetilde{\delta}_{k}^{(j)})}_2 \\
        \leq \norm{\bm{z}_1 - \bm{z}_2}_2 + \norm{\delta_{k}^{(j)}\rho_{\xi}(\bm{d}_1) - \widetilde{\delta}_k^{(j)} \rho_{\xi}(\bm{d}_2)}_2. \label{eqn:DR-CG-Net fiedality gradient update bound 1}
    \end{multline}
    As $\rho_{\xi}$ is 1-Lipschitz and bounded, in norm $\norm{\cdot}_2$, by $\xi$ then
    \begin{align}
        &\norm{\delta_k^{(j)}\rho_{\xi}(\bm{d}_1) - \widetilde{\delta}_k^{(j)}\rho_{\xi}(\bm{d}_2)}_2 \nonumber \\
        &= \norm{\delta_k^{(j)}\rho_{\xi}(\bm{d}_1) - \delta_k^{(j)}\rho_{\xi}(\bm{d}_2)+ \delta_k^{(j)}\rho_{\xi}(\bm{d}_2) - \widetilde{\delta}_k^{(j)}\rho_{\xi}(\bm{d}_2)}_2 \nonumber \\
        &\leq |\delta_k^{(j)}|\norm{\rho_{\xi}(\bm{d}_1) - \rho_{\xi}(\bm{d}_2)}_2 + \norm{\rho_{\xi}(\bm{d}_2)}_2\,\, |\delta_k^{(j)}-\widetilde{\delta}_k^{(j)}| \nonumber \\
        &\leq \delta \norm{\bm{d}_1 - \bm{d}_2}_2 + \xi |\delta_k^{(j)}-\widetilde{\delta}_k^{(j)}|. \label{eqn:DR-CG-Net Gradient bound}
    \end{align}
    Now, combining Lemma \ref{lemma:Lipschitz bound on Au(Auz-y)} with (\ref{eqn:DR-CG-Net Gradient bound}), (\ref{eqn:DR-CG-Net fiedality gradient update bound 1}), and the fact that $\norm{\overline{\bm{y}}_p}_2\leq y_{\max}$ produces
\begin{multline}
    \norm{v_k^{(j)}(\bm{z}_1,\bm{u}_1; \delta_{k}^{(j)}) - v_k^{(j)}(\bm{z}_2,\bm{u}_2; \widetilde{\delta}_{k}^{(j)})}_2 \\
    \leq \left(1 + \delta  (z_{\infty} p_{\max} y_{\max}  \norm{A}_2 \norm{A}_{\infty})^2\right) \norm{\bm{z}_1 - \bm{z}_2}_2  \\
    + r_2 \norm{\bm{u}_1 - \bm{u}_2}_2 + \xi |\delta_k^{(j)}-\widetilde{\delta}_k^{(j)}|. \label{eqn:DR-CG-Net fiedality gradient update bound 2}
\end{multline}

As $W_{k,d}^{(j)}\in\Omega_d$ and $\widetilde{W}_{k,d}^{(j)} \in\Omega_{d}$ for $d \in \mathbb{N}[L_c]$, then $W_{k,d}^{(j)}$ and $\widetilde{W}_{k,d}^{(j)}$ are defined by $f_{d-1}f_d k_d^2$ parameters and satisfy $\norm{W_{k,d}^{(j)}}_2\leq w_d$ and $\norm{\widetilde{W}_{k,d}^{(j)}}_2\leq w_d$. Additionally, $\delta_{k}^{(j)}\in \Omega_{L_c+1}$ is a positive real number satisfying $|\delta_k^{(j)}|\leq \delta$. Hence, the dimension of the parameter spaces, $\alpha_{d}$, and bounds on the parameter spaces, $\omega_{d}$, are given by (\ref{eqn:alpha, omega}). 

As the ReLU activation function is 1-Lipschitz and $||\bm{z}_1||_2\leq \sqrt{n}z_{\infty}$ then using Lemma \ref{lemma:lipschitz of fully-connected network}
\begin{multline}
    \norm{\mathcal{V}_k^{(j)}\left(\bm{z}_1\right) - \widetilde{\mathcal{V}}_k^{(j)}\left(\bm{z}_2\right)}_2 \leq \prod_{\ell = 1}^{L_c} w_{\ell} \norm{\bm{z}_1 - \bm{z}_2}_2 \\
    + \sum_{\ell = 1}^{L_c} \bigg(\sqrt{n}z_{\infty}\scalebox{1}{$\prod_{\substack{\ell' = 1 \\ \ell'\neq \ell}}^{L_c}$} w_{\ell'}\bigg) \norm{W_{k,\ell}^{(j)} -\widetilde{W}_{k,\ell}^{(j)}}_2. \label{eqn:DR-CG-Net subnet Lipschitz bound}
\end{multline}
Combining (\ref{eqn:DR-CG-Net subnet Lipschitz bound}) and (\ref{eqn:DR-CG-Net fiedality gradient update bound 2}) with (\ref{eqn:scale mapping module bound 1 DR-CG-Net}) then Assumption \ref{assumption:scale update method} holds with $r_{k,1}^{(j-1)} = r_1$, $r_{k,2}^{(j-1)} = r_2$, and 
\begin{align*}
\resizebox{\columnwidth}{!}{${\displaystyle
    \left(\bm{\theta}_{k,d}^{(j)}, r_{k,d,3}^{(j-1)}, \norm{\cdot}_{(d)}\right) = \begin{cases}
        \left(W_{k,d}^{(j)}, \sqrt{n} z_{\infty}\prod_{\substack{\ell = 1 \\ \ell\neq d}}^{L_c}w_{\ell}, \norm{\cdot}_2\right) & d = 1, \ldots, L_c \\
        (\delta_k^{(j)}, \xi, |\cdot|) & d = L_c+1.
    \end{cases}
    }$}
\end{align*}

Hence, by Proposition \ref{prop:scale variable lipschitz 1}, ${\displaystyle \hspace{.5cm} \widehat{r}_{k,1}^{(J)} = \prod_{j = 1}^{J} r_{k,1}^{(j-1)} = r_1^J, }$
\begin{align*}
    \widehat{r}_{k,2}^{(J)} &= \sum_{j = 1}^{J} r_{k,2}^{(j-1)}\prod_{\ell = j}^{J-1} r_{k,1}^{(\ell)} = r_2 \sum_{j =1}^J r_1^{J-j} = r_2\frac{1-r_1^J}{1-r_1},
\end{align*}
and similarly $\widehat{r}_{k,d,3}^{(j,J)} = r_{k,d,3}^{(j-1)}\prod_{\ell = j}^{J-1} r_{k,1}^{(\ell)} = r_{k,d,3}^{(j-1)} r_1^{J-j}$ is given as in (\ref{eqn:r_3 DR-CG-Net}). Therefore, applying Theorem~\ref{thm:GEB G-CG-Net} produces the GEB of DR-CG-Net in Theorem~\ref{thm:GEB DR-CG-Net}.
\end{proof}

\bibliographystyle{IEEEtran}
\bibliography{main}

\end{document}